\documentclass[runningheads]{llncs}

\usepackage{eccv}

\usepackage{eccvabbrv}

\usepackage{graphicx}
\usepackage{booktabs}

\newcommand{\citet}{\cite}
\pretocmd{\eqref}{Eq.~}{}{}
\newcommand{\jh}{}

\usepackage{etoolbox}
\usepackage{capt-of}
\usepackage{xcolor}
\usepackage{amsmath}
\usepackage{colortbl}
\usepackage{wrapfig}
\usepackage{multirow}

\usepackage{amssymb}
\usepackage{pifont}

\definecolor{JSViolet}{RGB}{71,15,244}
\definecolor{JSRed}{RGB}{205,44,78}
\newcommand{\cmark}{\textcolor{JSViolet}{\ding{51}}}
\newcommand{\xmark}{\textcolor{JSRed}{\ding{55}}}

\usepackage{algorithm}
\usepackage{algpseudocode}
\newcommand{\algrule}[1][.3pt]{\par\vskip.2\baselineskip\hrule height #1\par\vskip.2\baselineskip}

\usepackage[accsupp]{axessibility}  

\usepackage{hyperref}
\definecolor{cornellred}{rgb}{0.7, 0.11, 0.11}
\definecolor{navyblue}{rgb}{0,0.4,0.8}
\hypersetup{
  linkcolor = cornellred,
  citecolor  = navyblue,
  urlcolor = navyblue,
  colorlinks = true,
  breaklinks = true,
}
\usepackage{orcidlink}

\begin{document}

\title{Adversarial Robustification via \\ Text-to-Image Diffusion Models} 

\author{Daewon Choi\inst{1}\thanks{Equal contribution.}\orcidlink{0009-0003-6126-4675} \and
Jongheon Jeong\inst{2}\textsuperscript{$\star$}\orcidlink{0000-0002-4058-5774} \and
Huiwon Jang\inst{1} \and
Jinwoo Shin\inst{1}}
\authorrunning{D.~Choi et al.}

\institute{Korea Advanced Institute of Science and Technology (KAIST) \and Korea University \\
\email{\{daeone0920,huiwoen0516,jinwoos\}@kaist.ac.kr} \quad \email{jonghj@korea.ac.kr}}


\maketitle

\begin{abstract}
Adversarial robustness has been conventionally believed as a challenging property to encode for neural networks, requiring plenty of training data. In the recent paradigm of adopting off-the-shelf models, however, access to their training data is often infeasible or not practical, while most of such models are not originally trained concerning adversarial robustness. In this paper, we develop a scalable and model-agnostic solution to achieve adversarial robustness without using any data. Our intuition is to view recent text-to-image diffusion models as ``adaptable'' denoisers that can be optimized to specify target tasks. Based on this, we propose: (a) to initiate a denoise-and-classify pipeline that offers provable guarantees against adversarial attacks, and (b) to leverage a few synthetic reference images generated from the text-to-image model that enables novel adaptation schemes. Our experiments show that our data-free scheme applied to the pre-trained CLIP {could improve the (provable) adversarial robustness of its diverse zero-shot classification derivatives (while maintaining their accuracy), significantly surpassing prior approaches that utilize the full training data. Not only for CLIP, we also demonstrate that our framework is easily applicable for robustifying other visual classifiers efficiently. Code is available at \url{https://github.com/ChoiDae1/robustify-T2I}.}
\end{abstract}
\section{Introduction}
\label{sec:intro}
Arguably, recent breakthroughs in deep learning have been largely driven by massive data and model scaling \cite{radford2021learning, sam, Brown2020gpt3, Singh2023maeprepretrain}, which enabled many unprecedented capabilities in computer vision \cite{radford2021learning,jia2021scaling, sam, saharia2022imagen, Ramesh2021dalle}.
The \emph{worst-case} behaviors of models at scale, however, have been relatively under-explored in the literature, despite their increasing practical relevance \cite{zhang2022towards,carlini2023aligned,zou2023universal,zhou2023advclip}. 
\emph{Adversarial robustness} \cite{szegedy2013intriguing,madry2018towards,pmlr-v80-athalye18a,carlini2019evaluating} is one of popular objectives in this context: specifically, it aims to build a model that makes consistent predictions for \emph{every} input perturbation within a small, often imperceptible, bound. 
Although it has been demonstrated that many types of (``natural'') robustness can benefit from proper data scaling \cite{radford2021learning,wortsman2022robust,fang2022data}, \eg, combined with recent \emph{vision-language models} \cite{radford2021learning,jia2021scaling,Li2022blip,Singh2022flava}, 
the trend does not seem to hold for adversarial robustness so far \cite{mao2023understanding,carlini2023aligned}, particularly highlighting its challenging nature. 

Despite being a desirable property, pursuing adversarial robustness in practice has been viewed as a costly design decision. This is possibly due to that most of existing techniques for adversarial robustness require a specialized, less-scalable training scheme \cite{madry2018towards,cohen2019certified,wang2021beta} from enough data \cite{schmidt2018adversarially,rebuffi2021data}, even followed by significant performance trade-offs \cite{tsipras2018robustness,zhang2019theoretically}. 
As a result, many of the existing pretrained, off-the-shelf models widely used in the community remain susceptible to adversarial attacks \cite{lan2022alphazero,carlini2023aligned,zou2023universal}.
Provided that the specific training data used for such off-the-shelf models is frequently not publicly accessible, it becomes increasingly challenging for end users to secure adversarial robustness for these models, \eg, to incorporate them into security-concerned applications \cite{caruana2015intelligible,yurtsever2020survey}.

\begin{figure*}[tb]
\centering
\begin{subfigure}{0.475\textwidth}
    \centering
\includegraphics[width=\linewidth]{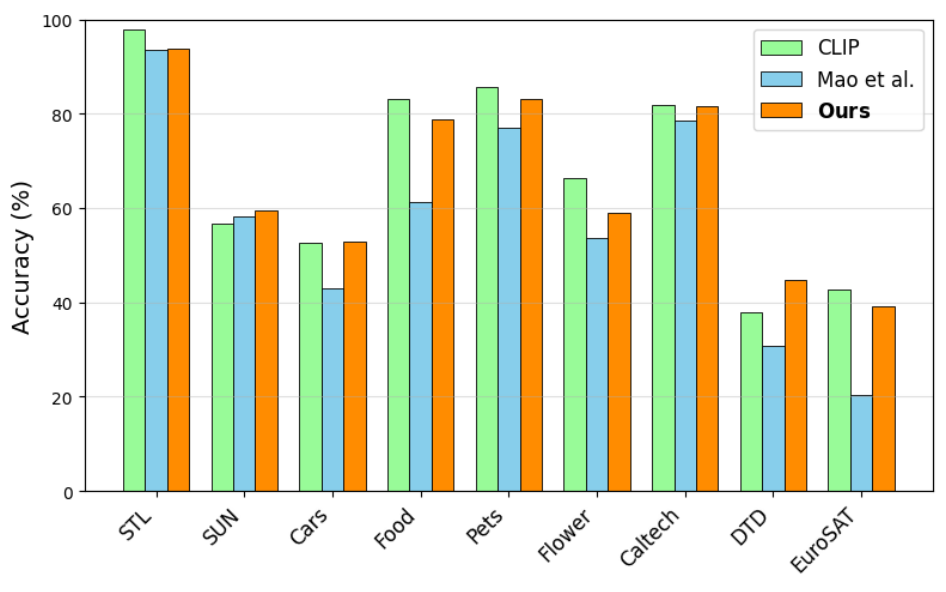}
    \caption{Clean accuracy}\label{fig:clean-chart}
\end{subfigure}
\hfill
\begin{subfigure}{0.475\textwidth}
    \centering
\includegraphics[width=\linewidth]{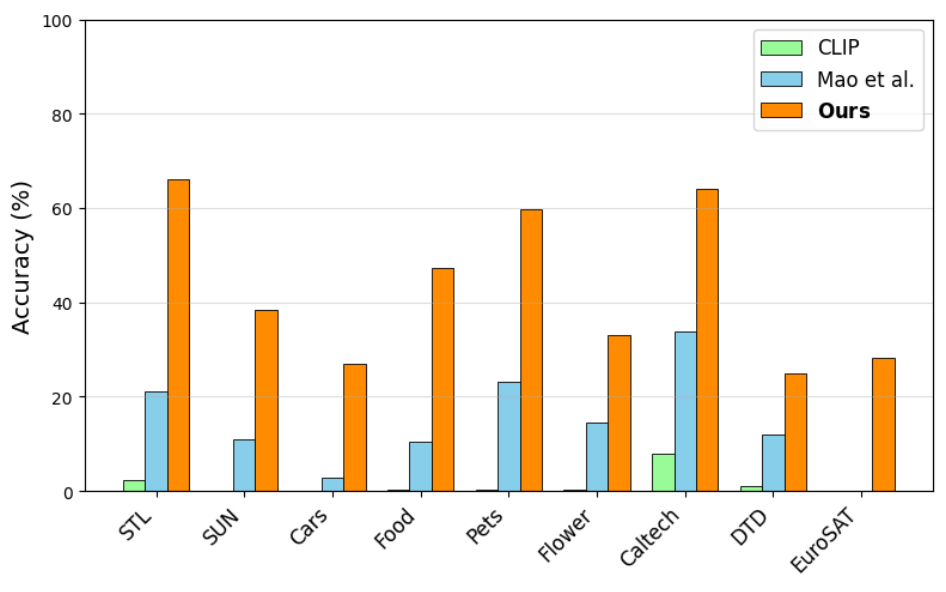}
    \caption{Robust accuracy}\label{fig:robust-chart}
\end{subfigure}
\caption{
{Comparison of clean and robust ($\|\varepsilon\|_2 \le 1.0$) accuracy on zero-shot classification: our framework (a) not only maintains the original accuracy of CLIP \cite{radford2021learning}; but also (b) significantly improves its robust accuracy, \eg compared to Mao et al.~\cite{mao2023understanding}.}}
\label{fig:clean-robust-chart}
\vspace{-0.2in}
\end{figure*}

A recent work \cite{mao2023understanding} has attempted to address the challenge through an adversarial contrastive fine-tuning scheme on CLIP \cite{radford2021learning} using external data such as ImageNet, in order to transfer the obtained robustness to other zero-shot downstream tasks. 
Yet, the method is limited in a sense that: (a) it is applicable only for vision-language models, (b) requiring a substantial amount of training data, \eg, as large as ImageNet in scale, to ensure its effectiveness. Furthermore, we observe that the approach is susceptible to an \emph{overfitting} to the fine-tuning data: \eg, the accuracy on downstream tasks often degrades after the fine-tuning (see \cref{fig:clean-chart}), and fails in transferring the robustness on datasets that significantly varies from ImageNet (see \cref{table:cd-robust-clean}).

\begin{figure*}[tb]
\centering
\begin{subfigure}{0.55\textwidth}
    \centering
    \includegraphics[width=\linewidth]{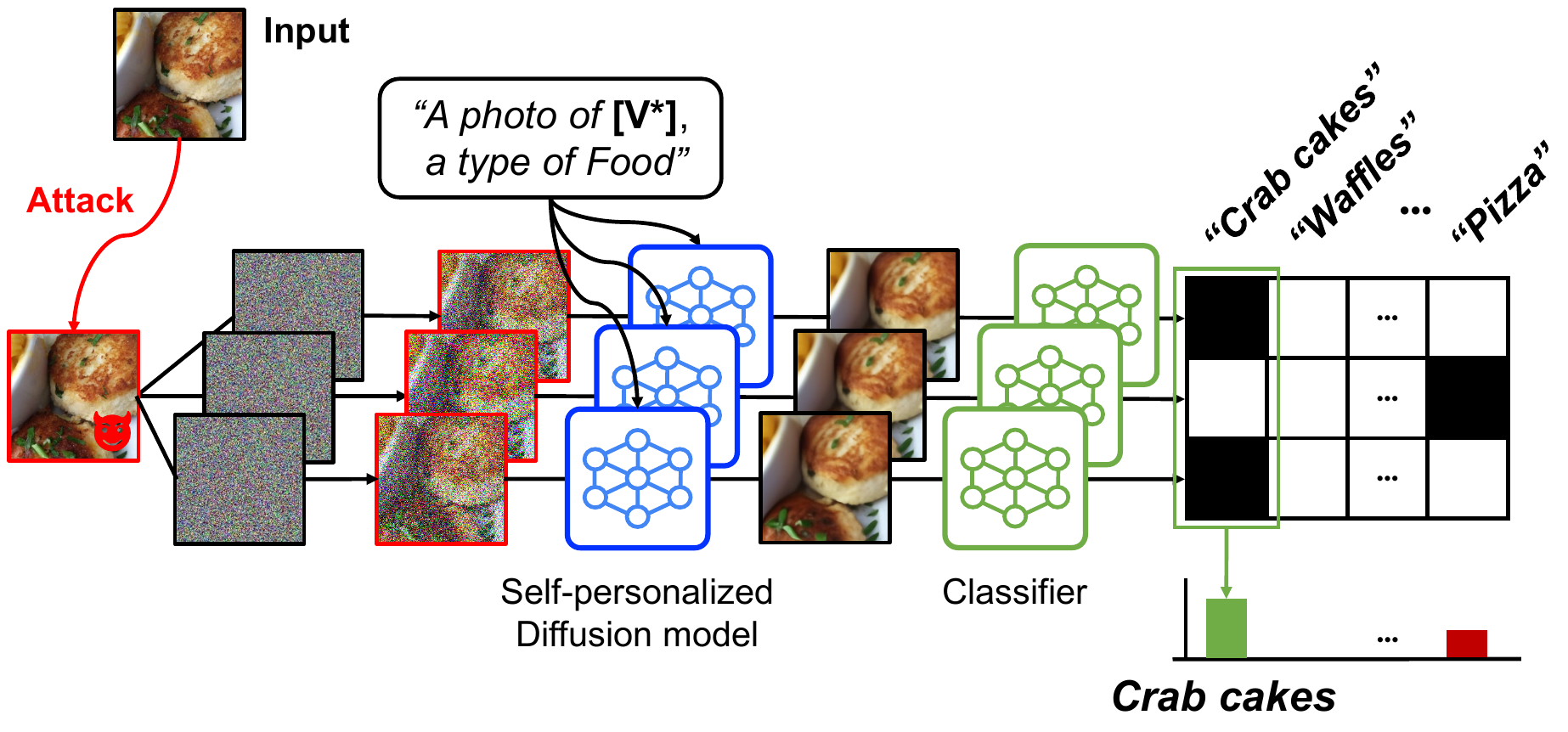}
 \caption{Denoised smoothing from text-to-image models}
    \label{fig:overview_randomized_smoothing}
\end{subfigure}
\hfill
\begin{subfigure}{0.42\textwidth}
    \centering
 \includegraphics[width=\linewidth]{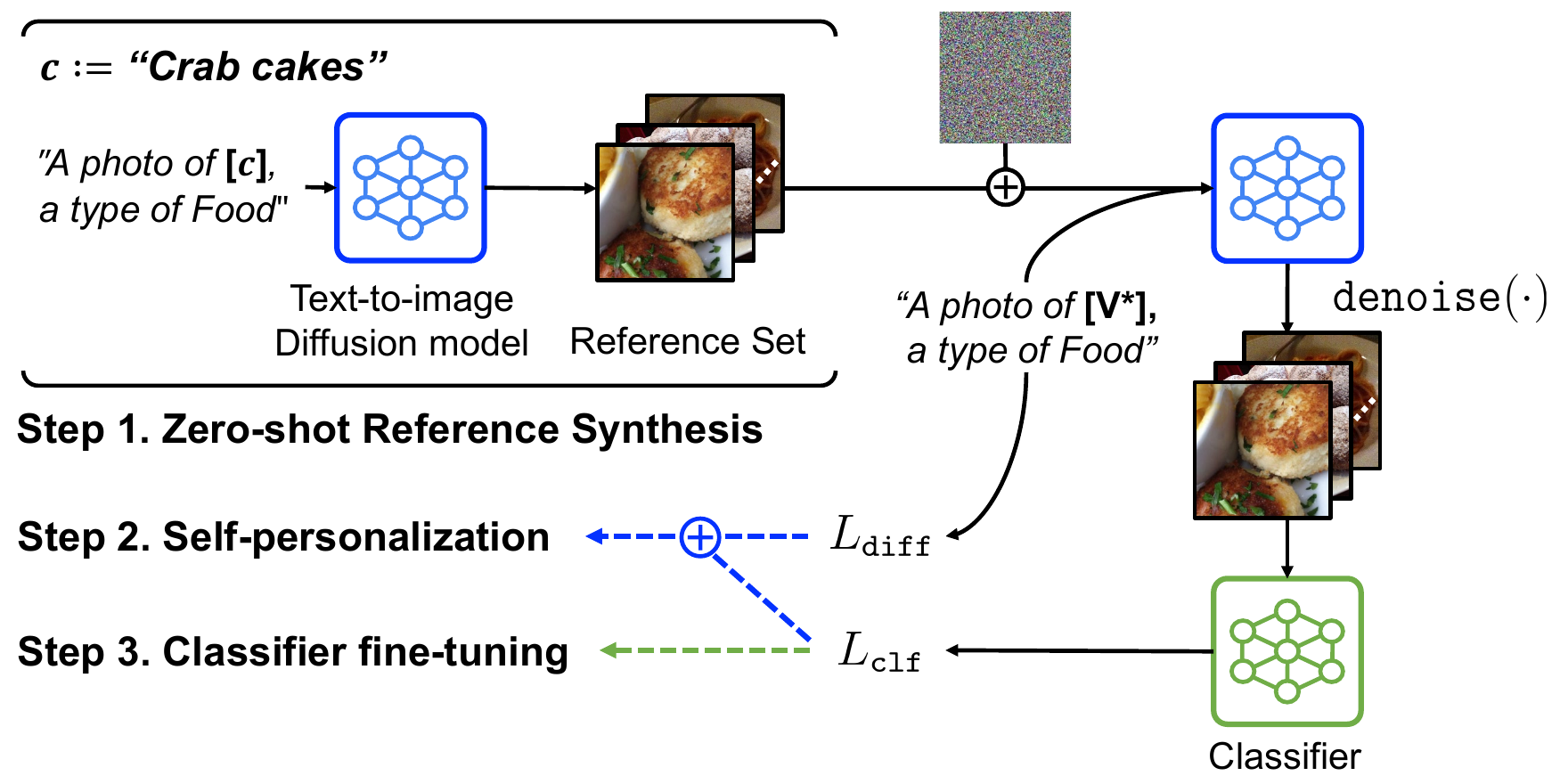}
    \caption{Self-adaptation schemes}   \label{fig:overview_framework_detail}
\end{subfigure}
\caption{An overview of the proposed framework: (a) during inference, we perform denoised smoothing with a self-personalized text-to-image diffusion model, having provable guarantees on adversarial robustness (\cref{sec:method:denoise-text-image-model}); (b) by utilizing synthetic references from the text-to-image  model, one can adapt both diffusion model and classifier for robustness (\cref{sec:method:zero-shot-adaptation}).}
\label{fig:method_overview}
 \vspace{-0.1in}
\end{figure*}

\vspace{0.05in}
\noindent
\textbf{Contribution. } 
In this paper, by leveraging recent text-to-image diffusion models, we propose a scalable framework {that does not require any external datasets in \emph{robustifying} image classifiers}.
Our framework is based on \emph{denoised smoothing} \cite{salman2020denoised}, a recent technique that constructs a provably-robust classifier from neural networks (\ie, \emph{certified defense} \cite{wong2018provable,cohen2019certified,wang2021beta,li2023sok}) through a ``denoise-and-classify'' pipeline, with a denoiser model on top of classifier.
Previous works upon denoised smoothing have only considered denoiser models that are optimized for the target classification task \cite{carlini2023diffusion,xiao2022densepure,jeong2023multiscale}, fully utilizing their training data. Here, we aim to extend its capability up to a next level, {\ie, robustifying without using data.
To this end, we utilize \emph{text-to-image super-resolution diffusion models} 
(like Imagen \cite{saharia2022imagen} or DeepFloyd-IF) with careful details we found, \eg, on setting proper diffusion timestep.
Upon the pipeline, we propose to utilize a few synthetic reference samples generated from the text-to-image model {based on the textual labels of target task}, 
and develop two adaptation schemes that boost the robustness: \textit{viz.}, fine-tuning (a) of diffusion model via classifier-guided personalization, and (b) of classifier from denoised samples {(see \cref{fig:method_overview})}. 

We conduct a comprehensive series of experiments to verify the effectiveness of our proposed framework, focusing on modern scenarios of classification using off-the-shelf models} and a wide spectrum of datasets \cite{dataset/ilsvrc,krause20133d,bossard14food,helber2019_eurosat}.
{First}, we show that our method could significantly enhance CLIP with a new robustness-accuracy frontier, even outperforming the previous method \cite{mao2023understanding} that utilizes the full ImageNet training data to robustify CLIP {(see Fig.~\ref{fig:robust-chart})}.
{In particular}, the effectiveness of our method is ``provably'' confirmed with the \emph{certified} robust accuracy, which guarantees the lower-bound of empirical robust accuracy.
{The robustness gains from our framework are shown to be much more consistent across datasets compared to the prior work \cite{mao2023understanding} that has been struggling to offer robustness when the target dataset significantly varies from those for fine-tuning, \eg, EuroSAT upon ImageNet. 
Next, we verify that our proposed framework can also be effective in robustifying other generic vision classifiers, \eg, an ImageNet pre-trained ResNet-50, even surpassing standard approaches to obtain adversarial robustness such as adversarial training \cite{madry2018towards}.} 

\vspace{0.05in}
\noindent
To summarize, we make the following contributions:
\begin{enumerate}
\itemsep 0.5em
\item To the best of our knowledge, {our framework is the first approach toward robustifying any given (off-the-shelf) vision classifiers} {without using any data.}
We utilize recent text-to-image diffusion models as denoisers that can be adapted to target tasks via text-conditioning, and show that incorporating them into the inference of pre-trained classifiers can be a scalable approach.

\item We further propose to generate a few reference samples re-utilizing the text-to-image diffusion model, and leverage them to adapt for individual tasks. We show that fine-tuning both the text-to-image diffusion model as well as classifier can be jointly beneficial to boost adversarial robustness.

\item We evaluate our framework in robustifying CLIP for a variety of zero-shot classification tasks: it could not only offer state-of-the-art robustness on the benchmark, but also show consistent gains across a wider range of datasets. We also verify the proposed framework offers robustness to generic vision classifiers other than CLIP, obtaining better robustness even compared to, \eg, the popular adversarial training from scratch.
\end{enumerate}

\section{Preliminaries}
\label{sec:prelim}

\textbf{Adversarial Robustness} \cite{szegedy2013intriguing,carlini2019evaluating} refers to a property of a classifier, say $f$, to make consistent prediction under \emph{every} possible perturbations $\delta$ within a boundary, \eg, an $\ell_2$-ball: it requires $f(x+\delta)=y$ for \emph{any} $\lVert \delta \rVert_2\leq\varepsilon$. In this respect, adversarial robustness of $f$ for an input $x$ can be measured by the \textit{minimum-distance} of adversarial perturbation \cite{carlini2019evaluating}:
\begin{align}\label{eq:min-distance}
R(x,y; f) := \operatorname{min}_{f(x^\prime)\neq y}\lVert x - x^\prime \rVert_2.
\end{align}
One of the key challenges to achieving adversarial robustness is the hardness of accurately measuring \eqref{eq:min-distance}, which has eventually falsified the robustness claim of many previous works in the literature \cite{pmlr-v80-athalye18a,tramer2020adaptive,uesato2018adversarial}.  

\vspace{0.05in}
\noindent
\textbf{Randomized Smoothing} \cite{lecuyer2019certified,cohen2019certified} 
is currently one of the state-of-the-art methods in obtaining provable guarantees on adversarial robustness from neural network-based classifiers. Given a classifier $f$ and an input $x$, randomized smoothing makes an inference by taking a majority vote of $f(x+\delta)$ for random Gaussian noise $\delta \sim \mathcal{N}(0,\,\sigma^{2}\textbf{I})$. Specifically, it defines a \textit{smoothed classifier} $\hat{f}$ as follows:
\begin{align}\label{eq:smoothed classifier}
\hat{f}(x) := \underset{c\in\mathcal{Y}}{\operatorname{argmax}}\ \mathbb{P}_{\delta\sim\mathcal{N}(0,\,\sigma^{2}\textbf{I})}[f(x+\delta) = c].
\end{align} 
Then, Cohen et al. \citet{cohen2019certified} have shown that the adversarial robustness of $\hat{f}$ at $x$ is guaranteed by a lower-bound $\underline{R}$:
\begin{multline}\label{eq:cohen-vertifed}
R(x,y; \hat{f}) \geq \sigma\cdot \Phi^{-1}(p_{\hat{f}}(x, y)) =: \underline{R},
\text{ where}\quad p_{\hat{f}}(x, y) := \mathbb{P}_{\delta}[f(x+\delta) = y],
\end{multline}
provided that $\hat{f}(x) = y$, otherwise $R(x,y; \hat{f}) := 0$. {Here, $\Phi$ denotes the standard Gaussian CDF.}
Remark from \eqref{eq:cohen-vertifed} that the \emph{certified robustness} $\underline{R}$ of $\hat{f}$ depends on $p_{\hat{f}}$, which is essentially the accuracy of $f$ at noisy inputs $x + \delta$.  

\vspace{0.05in}
\noindent
\textbf{Denoised Smoothing} \cite{salman2020denoised} is a recent framework for randomized smoothing that has enabled a more scalable design. Specifically, it constructs the base classifier $f$ for smoothing as a ``denoise-and-classify'' pipeline, which concatenates a Gaussian denoiser, say $\mathtt{denoise}(\cdot)$, with any standard classifier $f_{\mathtt{clf}}$ as follows:
\begin{align}\label{eq:carlini}
f(\hat{x}) := f_{\mathtt{clf}}(\mathtt{denoise}(\hat{x})).
\end{align}
Here, a ``good'' denoiser $\mathtt{denoise}(\cdot)$ should accurately reconstruct the semantic of $x$ from $\hat{x} := x+\delta$ with high probability of $\delta\sim\mathcal{N}(0,\,\sigma^{2}\textbf{I})$. 
In practice, this often requires the denoiser to be sufficiently optimized for the input distribution of target tasks, otherwise the performance of $\hat{f}$ is significantly limited by the denoiser \cite{salman2020denoised}. For tasks where an accurate denoiser is available, however, denoised smoothing can offer a strong design for randomized smoothing: \eg, Carlini et al.~\citet{carlini2023diffusion} achieved state-of-the-art certified robustness on ImageNet \cite{dataset/ilsvrc} by adopting a high-fidelity ImageNet diffusion model \cite{dhariwa2021diffusion} as the denoiser. The idea of leveraging diffusion models for ``denoise-and-classify'' has been also considered as an empirical defense \cite{yoon2021puri,nie2022DiffPure} (\ie, not certifiable): yet, such approaches also necessitate a separate diffusion model trained for the target dataset.

\vspace{0.05in}
\noindent
\textbf{Diffusion Model} \cite{ho2020denoising,nichol2021diffusion}
aims to learn a generative distribution $p_{\tt data}(x)$ by gradually denoising (or \textit{reverse} process) from noisy inputs from so-called \textit{diffusion} (or \textit{forward}) process. Formally, it first defines $x_t$ from a pure Gaussian noise $\varepsilon\sim\mathcal{N}(0,\textbf{I})$ and a \textit{timestep} $t\in[0, T]$, given $x$: 
\begin{align}\label{eq:forward pass}
x_t := \sqrt{\alpha_t}\cdot x + \sqrt{1-\alpha_t}\cdot\varepsilon,
\end{align}
where 
the factor $\alpha_t \in [0, 1]$ is a constant determined by $t$, which schedules the amount of noise typically as a monotonically decreasing function of $t$, 
\ie $x_t$ becomes noisier toward the unit Gaussian with increasing $t$. 
The key design of recent diffusion models is to parametrize a \emph{noise estimator} $\varepsilon_{\theta}(x_t, t)$, trained over $p_{\tt data}(x)$ and $t\in[0, T]$, which aims to predict the noise $\varepsilon$ added to $x_t$ given $t$.

\vspace{0.05in}
\noindent
\textbf{Text-to-Image Diffusion Model} \cite{rombach2022high, saharia2022imagen}
is a particular instance of diffusion models, which have recently demonstrated remarkable capabilities in generating high-fidelity images from natural language descriptions. Specifically, it considers modified architectures for noise estimator to condition on text, \ie, as the form of $\varepsilon_{\theta}(x_t, t, \tau_{\theta}(c))$, where $\tau_{\theta}$ is a \emph{text encoder} that maps a textual prompt $c$ into an embedding vector. Depending on specific designs, the latest architectures for text-to-image diffusion models roughly fall into two categories to handle high-resolution inputs: (a) \emph{latent diffusion models} \cite{rombach2022high}, which first map $x$ into a latent space of lower-resolution, and (b) \emph{cascaded diffusion models} \cite{ho2022cascaded,saharia2022imagen}, which train a lower-resolution diffusion model in the pixel space followed by multiple \emph{super-resolution} diffusion models of increasing resolutions.

\label{sec:related}
\section{{Text-to-Image Diffusion Models for Robustification}}
In this section, we introduce a scalable and model-agnostic framework to obtain adversarial robustness from image classifiers without accessing training data.
Given an image classifier $f: \mathcal{X} \rightarrow \mathcal{Y}$ trained on a data distribution $p_{\tt data}(x, y)$, we aim to construct a new classifier $\hat{f}$ that is adversarially-robust, without explicit knowledge of $p_{\tt data}$, \eg, for fine-tuning. The minimal assumption we make here is the textual knowledge of classes, \ie, $\mathcal{Y}:= \{c_i\}_{i=1}^K$ and each $c_i$ is given in text -- which is common in the recent literature of zero-shot classification \cite{radford2021learning, jia2021scaling, Li2022blip}. From this information, $\hat{f}$ is required to (a) maximize $\mathbb{E}_{(x, y)}[R(x, y; \hat{f})]$ \eqref{eq:min-distance}, while (b) minimizing the accuracy trade-off from robustifying $f$.

The key ingredient of our proposed framework to this end is the recent \emph{text-to-image diffusion models}, particularly those with \emph{pixel-level, cascaded diffusion models}, \eg, Imagen \cite{saharia2022imagen} or DeepFloyd-IF. 
Overall, our framework utilizes the model mainly in two different ways: (a) to be applied as the denoiser of the denoised smoothing pipeline \eqref{eq:carlini} (see \cref{sec:method:denoise-text-image-model}), and (b) to enable a fine-tuning of the classifier via personalization (see \cref{sec:method:zero-shot-adaptation}). 
The overall framework is illustrated in \cref{fig:method_overview}.    
\subsection{{Denoised Smoothing from Text-to-Image Diffusion Models}}\label{sec:method:denoise-text-image-model}

We first propose to utilize recent text-to-image diffusion models by means of their performance as \jh{a ``zero-shot'' denoiser}, so that it can be incorporated into the \textit{denoised smoothing} pipeline \eqref{eq:carlini}.   
Consider an input $x$ and its Gaussian perturbation, say $\hat{x} := x + \delta$, where $\delta\sim\mathcal{N}(0,\,\sigma^{2}\textbf{I})$ for a noise strength $\sigma$. Then, as observed by Carlini et al.~\citet{carlini2023diffusion}, one can correspond $\hat{x}$ into a \emph{timestep} of diffusion process \eqref{eq:forward pass}, \ie, to $x_{\hat{t}}$ for some $\hat{t}$. Specifically, it follows by:
\begin{align}\label{eq:relation-alpha}
    \sigma^2 = \frac{1-\alpha_{\hat{t}}}{\alpha_{\hat{t}}}.
\end{align}
With this relationship, one can search over the noise schedule $\alpha_t$ of a diffusion model for its corresponding timestep $\hat{t}$ given $\sigma$, which makes $\hat{x}$ compatible with the inherent denoiser of diffusion models, \eg, by scaling it with $\sqrt{\alpha_{\hat{t}}}$. 
 
Now, we introduce the detailed design components in adopting text-to-image models, of the general form of $\varepsilon_{\theta}(x_t, t, \tau_{\theta}(c))$, to define $\mathtt{denoise}(\cdot)$ in \eqref{eq:carlini}.

\vspace{0.05in}
\noindent
\textbf{Need for Pixel-based Diffusion Models. } Existing off-the-shelf text-to-image diffusion models are often based on \emph{latent diffusion model} architecture \cite{rombach2022high}, \eg, Stable Diffusion. Remark that, however, the pipeline of denoised smoothing \eqref{eq:carlini} requires a denoiser to directly denoise a given noisy input $\hat{x}$, which is in \emph{pixel-space}, making this kind of diffusion models incompatible for the pipeline. 
In this respect, our framework focuses on adopting \textit{cascaded diffusion models} into the pipeline, such as Imagen \cite{saharia2022imagen} and DeepFloyd-IF, another popular design choice for recent text-to-image models and those indeed consist of pixel-level diffusion models (of different resolutions).

\vspace{0.05in}
\noindent
\textbf{Super-resolution Diffusion Model as a Denoiser. }
More specifically, recall that cascaded diffusion models generally consist of (a) a low-resolution (\eg, $64\times64$) text-conditional diffusion model, followed by (b) multiple stages of {super-resolution} diffusion models (\eg, from $64\times64$ to $256 \times 256$) to enable higher-resolution generations in a scalable manner. 
Among these different diffusion models, our framework draws attention to the particular {attribute} of the \emph{super-resolution} models by means of {effective denoisers}. 
The choice is motivated by an intuition that super-resolution modules in cascaded diffusion models are more likely to be biased to ``reconstruct'' the original contents of $x$ given $\hat{x}$ in performing denoising, rather than generating new visual cues from scratch, which can be particularly beneficial in the pipeline of denoised smoothing. 

Formally, super-resolution diffusion models in cascaded designs typically parameterize a noise estimator as follows: 
\begin{equation}\label{eq:superres}
    \varepsilon_{\theta}(x_t, t, \tau_{\theta}(c) | \bar{x}_{t^\prime}, t^\prime),
\end{equation}
where $(x_t, t)$ is a noisy input and its timestep at the output resolution, and $\tau_{\theta}$ is a text encoder for conditioning. The additional condition compared to the standard models, \ie, $(\bar{x}_{t^\prime}, t^\prime)$, is from the previous (lower-resolution) module, processed by (a) first interpolating the output of the previous module up to the output resolution, followed by (b) mixing with Gaussian noise using a certain timestep $t^\prime$. 

\vspace{0.05in}
\noindent
\textbf{Timestep Correction. } 
In our context of adapting the model for denoised smoothing, we propose to set \emph{both inputs} of $x_t$ and $\bar{x}_{t^\prime}$ in \eqref{eq:superres} by $\sqrt{\alpha_{\hat{t}}}\cdot\hat{x}$, \ie, by $x_t = \bar{x}_{t^\prime} = \sqrt{\alpha_{\hat{t}}}\cdot\hat{x}$, where $\hat{t}$ is the timestep searched with respect to \eqref{eq:relation-alpha}. In this way, the super-resolution module $\varepsilon_{\theta}$ is ``self-conditioned'' by the information available from $\hat{x}$.
A surprisingly important detail to make this design work is on the timestep $t^\prime$: 
we find that setting a \emph{higher value} of timestep for $t^\prime$ than $\hat{t}$, despite being $x_t = \bar{x}_{t^\prime}$, is crucial for the denoising performance of the model. 
Specifically, we consider a \emph{correction factor} $k > 1$ as a hyperparameter to scale $t^\prime$, and propose to set:
\begin{equation}\label{eq:timestep_scale}
    t^\prime := k \cdot \hat{t}.
\end{equation}
This interesting behavior, specific to super-resolution diffusion models, can be explained by considering that $\bar{x}_{t^\prime}$ in \eqref{eq:superres} is originally assumed to be ``upsampled'' before applying a noise. Therefore, $\bar{x}_{t^\prime}$ is naturally expected to consist of narrower range of spatial frequencies, whereas the input given, $\sqrt{\alpha_{\hat{t}}} \cdot \hat{x}$, is directly from higher resolution: using higher values for $t^\prime$ is an effective way to reduce such excessive frequency information present in $\hat{x}$, given that it corresponds to an increased blurring in the denoising process.

\vspace{0.05in}
\noindent
\textbf{Overall Pipeline. } 
Putting together, our proposed {denoised smoothing based pipeline} is obtained using a text-conditional, super-resolution diffusion model $\varepsilon_\theta$. Specifically, given a noisy input $\hat{x} := x + \delta$, where $\delta \sim \mathcal{N}(0, \sigma^2 \textbf{I})$, we define a denoiser function for \eqref{eq:carlini} as follows:
\begin{equation}\label{eq:one-step-denoising-super}
\mathtt{denoise}_{\theta}(\hat{x}) :=
\hat{x} - \sigma \cdot  \varepsilon_{\theta}(\sqrt{\alpha_{\hat{t}}}\hat{x}, \hat{t}, \tau_{\theta}(\mathtt{C}(\text{`` ''})) | 
\sqrt{\alpha_{\hat{t}}}\hat{x}, k\hat{t}).
\end{equation}
Here, $\mathtt{C}(c)$ is a pre-defined textual ``template'' that implants a given (textual) label $c$, specific per task. For example, we use $\mathtt{C}(c) :=``\textit{A photo of a \{c\}, a type of food.}"$ for the Food dataset \cite{bossard14food} in our experiments, following Radford at al.~\citet{radford2021learning}. For the case of \eqref{eq:one-step-denoising-super}, which considers a zero-shot case that the label is not given, we simply put the \emph{empty string} \text{`` ''} for $c$. 
Once we have a concrete $\mathtt{denoise}(\cdot)$ at hand, as \eqref{eq:one-step-denoising-super}, any classifier $f$ that combines $\mathtt{denoise}(\cdot)$ can now be robustified via randomized smoothing \eqref{eq:smoothed classifier}. The \emph{smoothed classifier} it returns, $\hat{f}$, is provably robust within the certified radius it guarantees by \eqref{eq:cohen-vertifed} for each $x$. In practice, the overall smoothing procedure is statistically estimated with $n$ \textit{i.i.d.}~Gaussian noise from $\mathcal{N}(0, \sigma^2 \textbf{I})$: we provide the details for the estimation in \cref{supple:pred-cer}

\subsection{\jh{Self-adaptation Schemes}}\label{sec:method:zero-shot-adaptation}

Upon our framework introduced in \cref{sec:method:denoise-text-image-model},
we propose to re-utilize the text-to-image diffusion model to further improve its robustness. 
Consequently, we propose a two-step adaptation scheme of models, again using only the knowledge of textual label set $\mathcal{Y} = \{c_i\}_{i=1}^K$, \ie, \jh{without using concrete data in $\mathcal{X}$}.

\vspace{0.05in}
\noindent
\textbf{Reference Set Synthesis. }
We start by leveraging the text-to-image model to synthesize a few reference images from the textual labels. Concretely, for a given textual label $c\sim\mathcal{Y}$, we obtain its corresponding prompt $\mathtt{C}(c)$ and use it to generate a synthetic image $x^{g}$ by conditioning it into the text-to-image diffusion model. Repeating this process, we obtain high quality reference set $D^g = \{(x_i^{g}, c_i)\}_{i=1}^K$ only from the information of $\mathcal{Y}$.

\vspace{0.05in}
\noindent
\textbf{Classifier-Guided Self-personalization. }
For a given reference set $D^g$, we next perform a fine-tuning of the text-to-image diffusion model. 
We adopt \textit{DreamBooth} \cite{ruiz2023dreambooth} to this end, one of state-of-the-art method for personalizing text-to-image models. Specifically, it fine-tunes the given noise estimator network, $\varepsilon_{\theta}$, with a special prompt combining a \textit{unique identifier}, which is typically a list of meaningless characters (\eg, $``sks"$), to implant the information of $D^g$. 
After the personalization, one can now use $\mathtt{C}(``sks")$ in \eqref{eq:one-step-denoising-super} as a replacement of $\mathtt{C}(\text{`` ''})$ during its inference. 
Again, considering that $\varepsilon_{\theta}$ is a super-resolution diffusion model, we consider the following DreamBooth objective:
\begin{equation}\label{eq:dream-booth}
L_{\tt diff}(\theta)\\
:= \mathbb{E}_{x^g, \varepsilon, t}\left[||\varepsilon - \varepsilon_{\theta}(x^g_t, t, \tau_{\theta}(\mathtt{C}(``sks"))|x^g_t, kt)||^2_2\right],
\end{equation}
where $t\sim \mathcal{U}([0, T])$ is a random timestep, $\varepsilon\sim\mathcal{N}(0,\textbf{I})$ is Gaussian noise, and $\tau_{\theta}(\mathtt{C}(``sks"))$ is the textual embedding from $\mathtt{C}(``sks")$ through the (frozen) text encoder $\tau_{\theta}$. 

To further boost capability of text-to-image diffusion model by means of a denoiser model, particularly in the context of \emph{denoised smoothing}, we propose to regularize the model personalization objective, $L_{\tt diff}$, with the \emph{denoised classification loss}, namely as a \emph{classifier-guided} regularization $L_{\tt clf}$ of personalization. 
The regularization essentially simulates the ``denoise-and-classify'' pipeline of denoised smoothing. Specifically, for a given reference image $x^g$ in $D^g$,  
$x^g$ is first processed into a noisy image $x_{t}^{g}$ via \eqref{eq:forward pass} using random timestep $t\sim\mathcal{U}([0, T])$, followed by the (personalized version of) zero-shot denoising \eqref{eq:one-step-denoising-super}: obtaining a denoised image $\tilde{x}^{g} = \mathtt{denoise}_\theta(\frac{1}{\sqrt{\alpha_t}}\cdot x_t^{g})$. 
Therefore, given a classifier $f_\psi$, we propose to additionally minimize the following loss given the pair $(\tilde{x}^{g}, c)$:
\begin{equation}\label{eq:regularization-loss}
L_{\tt clf}(\theta, \psi) := \mathbb{E}_{(x^g, c)\sim D^{g}, t}\left[\mathbb{CE}(f_{\psi}(\tilde{x}^g), c)\right],
\end{equation}
where $\mathbb{CE}(\cdot, \cdot)$ is the cross-entropy loss. 
Overall, we minimize the following objective by combining the two losses:
\begin{align}\label{eq:final-person-loss}
\theta^* = \operatorname*{arg\,min}_{\theta}
\mathcal\{{L}_{\tt diff}(\theta) + \lambda \cdot L_{\tt clf}(\theta, \psi)\},
\end{align}
where $\lambda>0$ is a hyperparameter. The text encoder $\tau_{\theta}$ of the text-to-diffusion model and the classifier $f_\psi$ are fixed during the personalization.

\vspace{0.05in}
\noindent
\textbf{Classifier Fine-tuning. } Lastly, we also apply the denoised classification loss, $L_{\tt clf}$, to further optimize the classifier side: even after the personalization of the diffusion model using $D^g$ for more accurate denoising, the classifier may still be suboptimal due to the distribution mismatch between the clean and denoised images during denoised smoothing, as also suggested by Carlini et al.~\citet{carlini2023diffusion}. By also directly minimizing the loss for such ``denoise-and-classify'' images, one can further reduce the gap. Specifically, we minimize:
\begin{align}\label{eq:classifier-guided loss}
\psi^* = \operatorname*{arg\,min}_{\psi}
L_{\tt clf}(\theta^*, \psi).
\end{align}
Similarly, here we freeze the denoiser model $\theta^*$ during the optimization (of $\psi$).
\label{sec:method}

\section{Experiments}

We verify the effectiveness of our proposed framework focusing on its ability of improving adversarial robustness without using external data.
As far as we are aware, {our setup} has not been previously explored in the literature. 
For comparisons, we choose the following two recent methods as our closest baselines: (a) Mao et al.~\cite{mao2023understanding}, which fine-tune CLIP \cite{radford2021learning} on ImageNet for empirical adversarial robustness on other (zero-shot) classification tasks; and (b) Carlini et al.~\citet{carlini2023diffusion}, which consider an unconditional diffusion model optimized for target task in denoised smoothing, \eg, on ImageNet. We provide the experimental details, \eg, datasets, architectures, evaluation, fine-tuning, \etc, in \cref{supple:exp-detail}.

\subsection{Robustification of CLIP}\label{sec:exp:zero-shot}

Firstly, we evaluate our framework upon CLIP-B/32 \cite{radford2021learning} for mainly comparison with Mao et al. \citet{mao2023understanding}. Here, we not only consider standard zero-shot classification benchmarks, but also more domain-specialized datasets that significantly vary from ImageNet. These datasets are regarded as more challenging cases for Mao et al.~\cite{mao2023understanding} due to their reliance on ImageNet. We further evaluate the robustification performance of our framework on ImageNet compared to Mao et al.~\citet{mao2023understanding} and Carlini et al.~\citet{carlini2023diffusion} directly, both of which utilize the full ImageNet training data, contrary to ours that keeps the assumption of not using data.

\vspace{0.05in}
\noindent
\textbf{Results on Standard Zero-shot Benchmarks.} We evaluate the robustification performance of our proposed framework to CLIP-B/32 \cite{radford2021learning} covering an extensive zero-shot classification benchmark \cite{coates2011stl10,FeiFei2004caltech,bossard14food,Parkhi2012pets,Nilsback2008flower,Xiao2010sun,dataset/dtd, krause20133d}. Specifically, we compare how much our framework effective on improving the robust and clean accuracy of CLIP on these tasks without any data, considering two $\ell_2$-adversary of budget $\varepsilon\in\{0.5, 1.0\}$. 
We mainly compare with (a) Mao et al.~\citet{mao2023understanding}, an adversarial fine-tuning scheme using ImageNet, as well as with the performance of (b) the vanilla CLIP zero-shot classification and (c) CLIP-Smooth, which directly applies randomized smoothing (\eqref{eq:smoothed classifier}) to the CLIP model without using a denoiser.
 
\begin{table}[t]
\centering
\setlength{\tabcolsep}{5pt}
\caption{{Robust and clean accuracy (\%) on 8 zero-shot classification datasets using CLIP against $\ell_2$-adversary with $\varepsilon \in \{0.5, 1.0\}$. We additionally report certified accuracy at $\varepsilon$ for ``Ours'' in parentheses {(see ``Certified'')}. Bold indicates the best.}
 }\label{table:main-robust-clean}
\vspace{-0.2in}
\begin{subtable}{\columnwidth}
\centering
\caption{Robust accuracy (\%)}
\vspace{-0.05in}
\resizebox{0.9\textwidth}{!}{
\begin{tabular}{c|lcccccccc|c}
\toprule
&Method  & STL  & SUN & Cars & Food & Pets & Flower & DTD & Caltech & Average \\
\midrule
\multirow{5}{*}{\phantom{0}$\varepsilon=0.5$\phantom{0}}&
        CLIP& 10.8 & \phantom{0}1.2 & \phantom{0}0.0 & \phantom{0}1.8 & \phantom{0}2.7 &\phantom{0}0.8 & \phantom{0}2.7 & 12.0 & \phantom{0}4.0 \\
        &CLIP-Smooth& 42.6 & 23.7 & 14.3 &\phantom{0}8.9& 36.4 & 16.6 & 10.2 & 44.8 & 24.7 \\
        &Mao et al.~\cite{mao2023understanding}&59.4&29.9&12.5&32.9&51.2&33.5&18.8&56.2&36.8\\
        \cmidrule{2-11}
        \rowcolor[HTML]{EFEFEF}{\cellcolor{white}{}}&{\textbf{Ours}} & \textbf{80.4}& \textbf{41.8}  & \textbf{33.2}&\textbf{59.0} & \textbf{68.6}& \textbf{45.2}&\textbf{29.7}&\textbf{71.3}&\textbf{53.7}\\
        \rowcolor[HTML]{EFEFEF}{\cellcolor{white}{}}& \multicolumn{1}{r}{(Certified)} & (66.0) & (32.1) &(28.4) &(45.7) &(60.8)& (34.9) &(23.0)&(65.1)&(44.5)\\
        \midrule
        \multirow{5}{*}{\phantom{0}$\varepsilon=1.0$\phantom{0}}&
        CLIP&\phantom{0}2.4&\phantom{0}0.0&\phantom{0}0.0&\phantom{0}0.2 &\phantom{0}0.2 &\phantom{0}0.2&\phantom{0}1.0&\phantom{0}7.8&\phantom{0}1.5 \\
        &CLIP-Smooth&16.2&\phantom{0}5.8& \phantom{0}1.6&\phantom{0}0.4&\phantom{0}6.7&\phantom{0}4.5&\phantom{0}5.4&18.7&\phantom{0}7.4\\
        &Mao et al.~\cite{mao2023understanding}&21.2&11.0&\phantom{0}2.8&10.3&23.2&14.4&12.0&33.9&16.1 \\
        \cmidrule{2-11}
        \rowcolor[HTML]{EFEFEF}{\cellcolor{white}{}}& {\textbf{Ours}}& \textbf{66.0} & \textbf{38.3}&  \textbf{27.0}&\textbf{47.3}&\textbf{59.6} &\textbf{33.1} &\textbf{24.9}&\textbf{64.1}&\textbf{45.0}\\
        \rowcolor[HTML]{EFEFEF}{\cellcolor{white}{}}& \multicolumn{1}{r}{(Certified)} &(41.2)& (22.5)  &(18.9) &(28.9) & (46.5)&(18.9)&(18.2)&(55.8)&(31.4)\\
        \bottomrule
\end{tabular}}
\end{subtable}
\begin{subtable}{\columnwidth}
\centering
\vspace{0.1in}
\caption{Clean accuracy (\%)}
\vspace{-0.05in}
\hspace{0.025in}
\resizebox{0.9\textwidth}{!}{\begin{tabular}{c|lcccccccc|c}
        \toprule
        &Method  & STL  & SUN & Cars & Food & Pets & Flower & DTD & Caltech &
         Average \\
        \midrule
        -&CLIP&97.8&56.8&52.7&83.0&85.7&66.3&37.8&81.9&70.3\\
        \midrule
        \multirow{4.2}{*}{\phantom{0}$\varepsilon=0.5$\phantom{0}}
        &CLIP-Smooth& 75.0 & 46.8 & 42.1 & 52.3 & 66.7 & 43.5 & 17.2 & 68.3 & 51.5 \\
        &Mao et al.~\cite{mao2023understanding}&\textbf{94.8}&\textbf{60.0}&48.7&69.7&80.8&57.7&34.0&79.7&65.7\\
        \cmidrule{2-11}
        \rowcolor[HTML]{EFEFEF}{\cellcolor{white}{}}&{\textbf{Ours}} & \bf{94.8} & 58.6 & \bf{54.1} & \bf{80.2} & \bf{83.6} & \bf{61.4} & \bf{42.7} & \bf{81.7} & \bf{69.6} \\
        \rowcolor[HTML]{EFEFEF}{\cellcolor{white}{}}&\multicolumn{1}{r}{(Certified)} &(90.4) & (55.4) & (49.7)&(74.5) & (81.9)& (58.7)&(38.9)& (79.1)&(66.1) \\
        \midrule
        \multirow{4.2}{*}{\phantom{0}$\varepsilon=1.0$\phantom{0}}
        &CLIP-Smooth & 32.4 & 27.7 & 40.8 & 31.3 & 43.8 & 36.8 & \phantom{0}7.0 & 54.0 & 34.2 \\
&Mao et al.~\cite{mao2023understanding}&93.4&58.2&42.9&61.2&77.0&53.6&30.8&78.5&62.0 \\
        \cmidrule{2-11}
        \rowcolor[HTML]{EFEFEF}{\cellcolor{white}{}}&{\textbf{Ours}} & \bf{93.8} & \bf{59.4} & \bf{52.9} & \bf{78.8} & \bf{83.1} & \bf{58.9} & \bf{39.1} & \bf{81.7} & \bf{68.5} \\
        \rowcolor[HTML]{EFEFEF}{\cellcolor{white}{}}&\multicolumn{1}{r}{{(Certified)}} & (80.2) & (53.6)& (45.5)&(64.8)&(77.7)&(48.3) &(32.4)&(75.7)&(59.8)\\
        \bottomrule
       \end{tabular}}
\end{subtable}
\vspace{-0.2in}
\end{table}
In \cref{table:main-robust-clean}, we compare the robust and clean accuracy of our framework with the baselines, respectively.
We observe that the vanilla CLIP model is originally vulnerable to adversarial attacks: their average robust accuracy at $\varepsilon=0.5$ is decreased from the clean accuracy by $66.3\%$ ($70.3\%\rightarrow4.0\%$), near the chance level. Although \mbox{CLIP-Smooth} obtains better average robustness than vanilla CLIP ($4.0\%\rightarrow24.7\%$ at $\varepsilon=0.5$) through randomized smoothing, however, it is still susceptible to perturbations with a higher bound, \ie, $\varepsilon=1.0$. Mao et al.~\citet{mao2023understanding} fairly outperforms these baselines: nevertheless, we observe that it tends to exhibit {insufficient robustness gains for ``domain-specific'' datasets, \eg, Cars and DTD, as further confirmed in \cref{table:cd-robust-clean}.}
\begin{table}[t]
\centering
\setlength{\tabcolsep}{5pt}
\caption{{Robust and clean accuracy (\%) on three domain-specialized benchmarks using CLIP against $\ell_2$-adversary with $\varepsilon \in \{0.5, 1.0\}$. We report certified accuracy in parentheses. Bold indicates the best and runner-up is underlined.}
 }\label{table:cd-robust-clean}
\vspace{-0.1in}
\resizebox{0.8\textwidth}{!}{
\begin{tabular}{clccc|ccc}
\toprule
\multirow{3.8}{*}{}& 
\multirow{3.8}{*}{Method}& \multicolumn{3}{c}{Robust accuracy (\%)} & \multicolumn{3}{c}{Clean accuracy (\%)}\\
\cmidrule{3-8}
&&CropDisease&EuroSAT&ISIC&CropDisease&EuroSAT&ISIC\\
\midrule
\multirow{5}{*}{\
\phantom{0}$\varepsilon=0.5 $\phantom{0}}&CLIP&\phantom{0}0.0&\phantom{0}0.0&\phantom{0}0.0&\textbf{20.9}&\underline{42.6}&\underline{27.3}\\
&CLIP-Smooth&\phantom{0}1.8&\underline{13.6}&\phantom{0}\underline{9.0}&\phantom{0}7.1&16.6&14.3\\
&Mao et al.~\cite{mao2023understanding}&\phantom{0}\underline{2.2}&\phantom{0}0.6&\phantom{0}4.2&16.0&25.0&26.9\\
\cmidrule{2-8}
\rowcolor[HTML]{EFEFEF}{\cellcolor{white}{}}&\textbf{Ours}&\textbf{11.5}&\textbf{29.0}&\textbf{17.6}&\underline{20.3}&\textbf{45.2}&\textbf{35.7}\\
\rowcolor[HTML]{EFEFEF}{\cellcolor{white}{}}& \multicolumn{1}{r}{(Certified)}&\phantom{0}(4.3)&(11.0)&\phantom{0}(5.8)&(16.8)&(39.0)&(33.7)\\
\midrule
\multirow{5}{*}{\
\phantom{0}$\varepsilon=1.0 $\phantom{0}}&CLIP&\phantom{0}0.0&\phantom{0}0.0&\phantom{0}0.0&\textbf{20.9}&\underline{42.6}&\textbf{27.3}\\
&CLIP-Smooth&\phantom{0}\underline{1.2}&\underline{11.8}&\phantom{0}\underline{4.6}&\phantom{0}4.3&18.2&10.6\\
&Mao et al.~\cite{mao2023understanding}&\phantom{0}0.6&\phantom{0}0.0&\phantom{0}2.2&\underline{16.0}&20.4&22.0\\
\cmidrule{2-8}
\rowcolor[HTML]{EFEFEF}{\cellcolor{white}{}}&\textbf{Ours}&\phantom{0}\textbf{4.9}&\textbf{28.2}&\phantom{0}\textbf{8.4}&15.8&\textbf{44.8}&\underline{26.3}\\
\rowcolor[HTML]{EFEFEF}{\cellcolor{white}{}}&\multicolumn{1}{r}{(Certified)}&\phantom{0}(0.8)&\phantom{0}(5.0)&\phantom{0}(1.4)&\phantom{0}(8.4)&(37.8)&(22.0)\\
\bottomrule
\end{tabular}}
\vspace{-0.05in}
\end{table}

Our framework shows a significant improvement in robustness over other baselines across entire datasets and $\varepsilon$. For instance, we obtain $16.9\%$ average robust accuracy gain at $\varepsilon=0.5$ compared to Mao et al.~\citet{mao2023understanding} and this discrepancy becomes more larger ($16.9\%\rightarrow28.9\%$) as $\varepsilon$ is larger ($0.5\rightarrow1.0$). Moreover, the certified robust accuracy of our framework also outperforms the (empirical) robust accuracy of other baselines across all datasets and $\varepsilon$, \ie, the ``lower-bound'' robust accuracy already outperforms the empirical robustness of other baselines. 

These all results show strong adversarial robustness without external data is possible via our framework. Moreover, both the average empirical and certified clean accuracy of our framework not only surpass other baselines but also outperform the clean accuracy of vanilla CLIP in some datasets, such as Cars ($52.7\%\rightarrow54.1\%$), DTD ($37.8\%\rightarrow42.7\%$). These results indicate the flexibility of our framework, which is able to trade-off between robust and clean accuracy.

\vspace{0.05in}
\noindent
\textbf{Results on Domain-Specific Benchmarks.} \jh{Next, we focus our evaluation on several domain-specific datasets as more challenging but practical scenarios, namely on CropDiseases \cite{mohanty2016crop}, EuroSAT \cite{helber2019_eurosat}, and ISIC \cite{codella2018isic}: \eg, EuroSAT consists of photos specifically taken from satellites. 
In \cref{table:cd-robust-clean}, we observe that Mao et al.~\cite{mao2023understanding} exhibits particularly bad, even near-zero, robustness on these datasets, possibly due to the model itself being fine-tuned only from a domain represented by ImageNet. Our framework still shows consistent robustness gains here: \eg, it offers a significantly higher robustness in EuroSAT at $\varepsilon=1.0$ of $28.2\%$. Similarly to \cref{table:main-robust-clean}, 
our framework also maintains the clean accuracy of CLIP, notably even surpassing it on EuroSAT and ISIC.}
\begin{table}[t]
\centering
\setlength{\tabcolsep}{5pt}
\caption{{Robust and clean accuracy (\%) on ImageNet using CLIP against $\ell_2$-adversary with $\varepsilon \in \{0.5, 1.0\}$. We report certified accuracy in parentheses. Bold indicates the best and runner-up is underlined.}}
\label{imagenet-table}
\vspace{-0.1in}
\resizebox{0.8\textwidth}{!}{
\begin{tabular}{lccc|cc}
\toprule
\multirow{3.8}{*}{Method}& \multirow{3.8}{*}{Data-free?} & \multicolumn{2}{c}{Robust accuracy (\%)} & \multicolumn{2}{c}{Clean accuracy (\%)}\\
\cmidrule{3-6}
&&$\varepsilon=0.5$ & $\varepsilon=1.0$ &$\varepsilon=0.5$ & $\varepsilon=1.0$ \\
\midrule
CLIP & \cmark &\phantom{0}1.4~\phantom{(00.0)} & \phantom{0}0.2 \phantom{(00.0)} & \textbf{58.2} \phantom{(00.0)} & \textbf{58.2} \phantom{(00.0)}\\
CLIP-Smooth & \cmark & 16.8 \phantom{0}(9.8) & \phantom{0}2.2 \phantom{0}(1.2) & 45.2 (25.0) & 35.2 \phantom{0}(3.8)\\
\rowcolor[HTML]{EFEFEF}{\textbf{Ours (w/o adapt)}} & \cmark & \underline{40.0} (29.6) & 31.0 (17.6) & 56.2 (50.8) & 55.2 (42.0)\\
\rowcolor[HTML]{EFEFEF}{\textbf{Ours}} & \cmark & \textbf{42.6} (\textbf{34.2}) & \underline{31.4} (\textbf{20.6}) & \underline{57.6} (\textbf{53.4}) & \underline{56.2} ({\textbf{46.0}})\\ 
\midrule
Mao et al.~\cite{mao2023understanding} & \xmark & 26.0 \phantom{(00.0)} & 12.3 \phantom{(00.0)} & 51.2 \phantom{(00.0)} & 47.2 \phantom{(00.0)}\\
Carlini et al.~\cite{carlini2023diffusion} & \xmark & 38.6 (30.2) & \textbf{32.4} (19.8) & 54.4 (49.8) & 53.6 (44.2)\\
\bottomrule
\end{tabular}}
\vspace{-0.15in}
\end{table}

\vspace{0.05in}
\noindent
\textbf{Results on ImageNet. } {Finally, we show that our robustification scheme (without using any data)} can be competitive and even better compared to those directly accessing to training data. We consider ImageNet \cite{dataset/ilsvrc} for the evaluation, considering that Mao et al.~\citet{mao2023understanding} fine-tunes directly on ImageNet. In addition to Mao et al.~\citet{mao2023understanding}, we also consider Carlini et al.~\citet{carlini2023diffusion} as another baseline, by considering a denoise-and-classify pipeline that combines CLIP with an unconditional ImageNet diffusion model \cite{dhariwa2021diffusion}: this can provide a clearer comparison on the effectiveness of our zero-shot denoised smoothing (\cref{sec:method:denoise-text-image-model}).  

In \cref{imagenet-table}, we report robust and clean accuracy, comparing our framework with these baselines. Even while Mao et al.~\citet{mao2023understanding} and Carlini et al.~\citet{carlini2023diffusion} directly access the ImageNet training data, our framework achieves better (or competitive) performance to them. Compared with Mao et al.~\citet{mao2023understanding}, we obtain not only $16.6\%$ and $19.1\%$ gains in empirical robust accuracy at $\varepsilon=0.5$ and $1.0$ but also $8.2\%$ and $8.3\%$ even from certified robust accuracy. 
Although Carlini et al.~\citet{carlini2023diffusion} gets a slightly better empirical robust accuracy at $\varepsilon=1.0$, our framework still achieves a higher certified robust accuracy. {Here, we notice that our proposed self-adaptation schemes (\cref{sec:method:zero-shot-adaptation}) play a crucial role for the gains: \eg, it contributes to a $4.6\%$ gain in certified accuracy ($29.6\%\rightarrow34.2\%$) at $\varepsilon=0.5$.}
Regarding the clean accuracy, our framework notably shows only a $0.6\%$ gap $\varepsilon=0.5$ compared to CLIP. 
All these results demonstrate the superiority of our framework in ensuring sufficient robust and clean accuracy, even compared with models directly accessing training data.

\begin{table}[t]
\centering
\setlength{\tabcolsep}{5pt}
\caption{{Robust and clean accuracy (\%) ResNet-50 on ImageNet across training schemes, using $\ell_2$-adversary with $\varepsilon \in \{0.5, 1.0\}$. We report certified accuracy in parentheses. Bold indicates the best and runner-up is underlined.}}\label{table:resnet-table}
\vspace{-0.1in}
\centering
\resizebox{0.9\textwidth}{!}{
\begin{tabular}{lccc|cc}
\toprule
\multirow{3.8}{*}{Method}& \multirow{3.8}{*}{Data-free?} & \multicolumn{2}{c}{Robust accuracy (\%)} & \multicolumn{2}{c}{Clean accuracy (\%)}\\
\cmidrule{3-6}
&&$\varepsilon=0.5$ & $\varepsilon=1.0$ &$\varepsilon=0.5$ & $\varepsilon=1.0$ \\
\midrule
Standard Training&\xmark& \phantom{0}5.2 \phantom{(00.0)} & \phantom{0}1.0 \phantom{(00.0)}& \textbf{74.4} \phantom{(00.0)} & \textbf{74.4} \phantom{(00.0)}\\
\rowcolor[HTML]{EFEFEF}{\textbf{$+$ Ours (w/o adapt)}}&\cmark&\underline{56.2} (47.0) & 44.2 (27.4)&\underline{73.0} (67.0) & 68.8 (57.2)\\
\rowcolor[HTML]{EFEFEF}{\textbf{$+$ Ours}}&\cmark&\textbf{57.0} (\textbf{50.4}) & \textbf{47.8} (34.0) &70.4 (\textbf{68.2}) & \underline{71.8} (\textbf{60.8})\\
\midrule
Adversarial Training~\cite{madry2018towards} &\xmark& 51.0 \phantom{(00.0)} & \underline{46.8} \phantom{(00.0)} & 55.0 \phantom{(00.0)} & 55.0 \phantom{(00.0)}\\
Randomized Smoothing~\cite{cohen2019certified} & 
\xmark&55.2 (48.6) & 43.8 (\textbf{37.0})&65.4 (66.8) & 55.4 (57.0)\\
Carlini et al.~\cite{carlini2023diffusion} & \xmark&\underline{56.2} (49.2) & 45.2 (33.2) &72.6 (67.4) & 70.0 (57.8)\\
\bottomrule
\end{tabular}}
\vspace{-0.15in}
\end{table}

\subsection{Robustification of Generic Vision Classifiers}\label{sec:exp:resnet}

{We further validate the scalability of our framework in robustifying classifiers other than CLIP, particularly considering ResNet-50 pre-trained on ImageNet. Here, we regard Carlini et al.~\cite{carlini2023diffusion} as a baseline again. For a thorough comparison, we consider two additional baselines: Adversarial Training~\cite{madry2018towards} and Randomized Smoothing~\cite{cohen2019certified}, both trained from scratch using the full ImageNet training data.}

\cref{table:resnet-table} reports the results in robust and clean accuracy on ImageNet.
Compared with the standard training, we obtain significant $51.8\%$ ($5.2\%\rightarrow57.0\%$ at $\varepsilon=0.5$) and $46.8\%$ ($1.0\%\rightarrow47.8\%$ at $\varepsilon=1.0$) gains in robust accuracy. These also outperform other baselines as well, \eg it surpasses Carlini et al.~\cite{carlini2023diffusion} in even certified robust accuracy. Again, the self-adaptation schemes contributes significantly to the gains, \eg by $6.6\%$ at $\varepsilon=1.0$.

\subsection{Ablation Study}\label{sec:exp:ablation}

We perform an ablation study to investigate the individual effectiveness of our framework. Here, we main compare the certified test accuracy of smoothed classifiers on ImageNet, 
as well as the \textit{average certified radius} (ACR; higher is better) \cite{zhai2020macer}, for a collective view of overall certified robustness. 

\begin{figure}[t]
\begin{minipage}{\textwidth}
\centering
\vspace{0.08in}
\begin{minipage}[c]{0.48\textwidth}
\centering
\includegraphics[width=\linewidth]{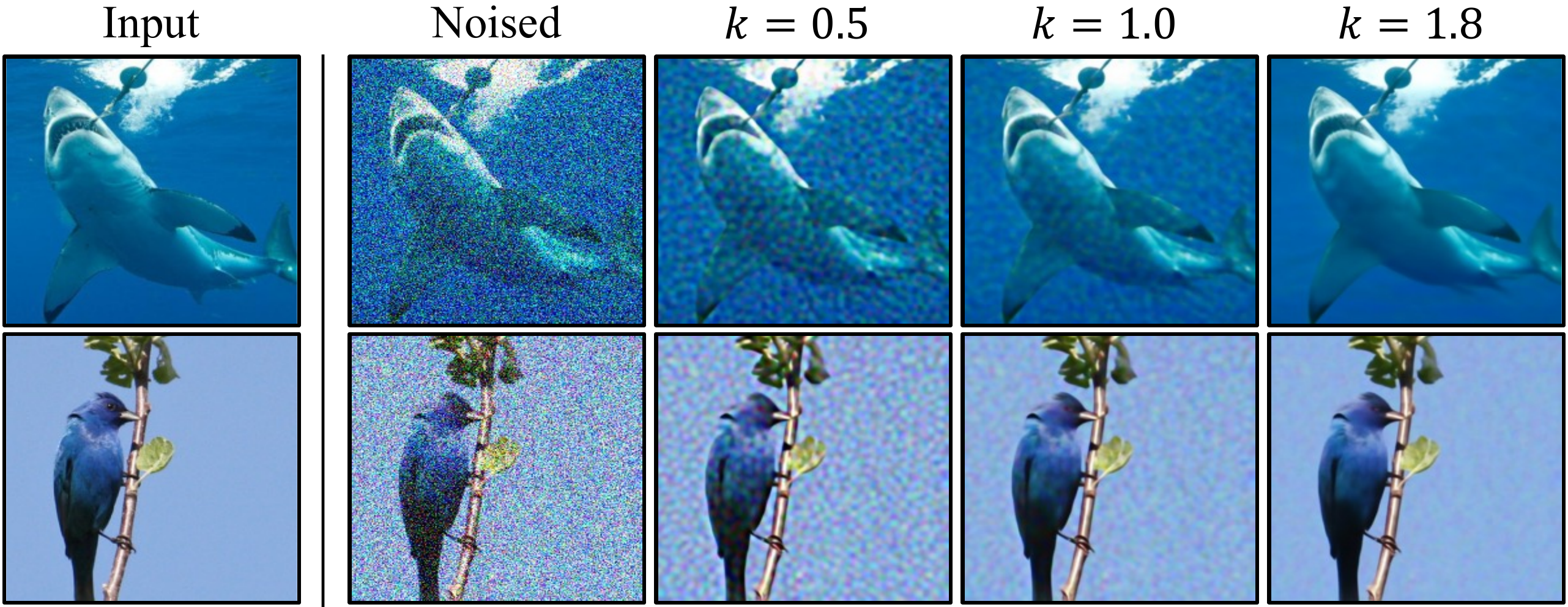}
\vspace{-0.25in}
\captionof{figure}{Qualitative comparisons of denoised images on varying correct factor $k$. 
We compared the denoised outputs from \eqref{eq:one-step-denoising-super} under Gaussian noise of $\sigma=0.25$.
}\label{figure:time_correct_factor}
\end{minipage}
\hspace{0.1in}
\begin{minipage}[c]{0.47\textwidth}
\centering
\setlength{\tabcolsep}{5pt}
\captionof{table}{
Ablation study on $k$ in \eqref{eq:timestep_scale}. Bold indicates the best.}\label{table:ablation-correct-factor}
\vspace{0.03in}
\resizebox{\linewidth}{!}{
\begin{tabular}[b]{ccccccccc}
\toprule
\multirow{2}{*}{}&&&\multicolumn{6}{c}{Certified accuracy at $\varepsilon$ (\%)}\\
\cmidrule{4-9}
$\sigma$&$k$&ACR&0.0&0.25&0.5&0.75&1.0&1.25\\
\cmidrule{1-1}\cmidrule(lr){2-2}\cmidrule(lr){3-3}\cmidrule{4-9}
\multirow{3}{*}{0.25}
&0.5&0.152&36.2&23.0&14.3&\phantom{0}6.6\\
&1.0&0.229&46.0&35.6&25.2&13.4\\
\rowcolor[HTML]{EFEFEF}{\cellcolor{white}{}} & \cellcolor{white}{1.8} &\textbf{0.277}&\textbf{50.8}&\textbf{41.8}&\textbf{29.6}&\textbf{19.2} & & \\
\midrule
\multirow{3}{*}{0.50}
&0.5&0.080&13.2&10.2&\phantom{0}6.0&\phantom{0}4.3&\phantom{0}3.4&\phantom{0}2.0\\
&1.0&0.200&28.8&22.0&16.4&11.2&\phantom{0}8.4&\phantom{0}6.0\\
\rowcolor[HTML]{EFEFEF}{\cellcolor{white}{}} & \cellcolor{white}{1.8} &  \textbf{0.367}&\textbf{42.0}&\textbf{35.4}&\textbf{29.2}&\textbf{23.8}&\textbf{17.6}&\textbf{13.0}\\
\bottomrule
\end{tabular}}
\end{minipage}
\end{minipage}
\vspace{-0.15in}
\end{figure}
\vspace{0.05in}
\noindent
\textbf{Timestep Correction. } We further analyze the influence of correction factor $k$ in \eqref{eq:timestep_scale} both qualitatively and quantitatively.
Specifically, we compare three different correction factors $k\in\{0.5, 1.0, 1.8\}$, including our default choice of $k=1.8$ in our experiments.\footnote{We chose $k$ by comparing the CLIP accuracy of denoised images from a small subset of ImageNet, as reported in 
\cref{supple:correct-factor}.} 
Here, $k=1.0$ means that $t^\prime$ is not corrected when used for \eqref{eq:one-step-denoising-super}. In \cref{figure:time_correct_factor}, we observe that using a higher value for correction factor $k$ leads to clearer denoised outputs. 
\cref{table:ablation-correct-factor} also confirms that the certified robustness denoised smoothing obtained is significantly impacted by the quality of the denoised samples. These results show that adjusting timestep $t^\prime$ is a critical design choice to make existing super-resolution diffusion models as a powerful denoiser model for denoised smoothing. 

\setlength{\intextsep}{0pt}
\begin{wraptable}{r}{0pt}
\centering
\setlength{\tabcolsep}{5pt}
\hspace{-0.05in}
\parbox[t]{0.4\textwidth}{
\vspace{-0.18in}
\caption{\jh{Ablation study on the proposed self-adaptation schemes. Bold and underline indicate the best and runner-up, respectively.}
}\label{table:ablation_component}
\resizebox{0.4\textwidth}{!}{
\begin{tabular}{ccccccc}
\toprule
\multirow{2}{*}{}&\multicolumn{2}{c}{Adapt.} & \multicolumn{4}{c}{Certified accuracy at $\varepsilon$ (\%)}\\
\cmidrule(lr){2-3} \cmidrule{4-7}
$\varepsilon$& T2I & CLIP &ImageNet&STL&SUN&Food\\
\cmidrule{1-1}\cmidrule(lr){2-3}\cmidrule{4-7}
\multirow{3}{*}{0.5}&\xmark&\xmark&29.6&55.2&28.3&43.6\\
&\cmark&\xmark&\underline{31.8}&\textbf{66.0}&\underline{30.7}&\underline{43.8}\\
\rowcolor[HTML]{EFEFEF}{\cellcolor{white}{}} & \cellcolor{white}{\cmark} & \cellcolor{white}{\cmark} &\textbf{34.2}&\textbf{66.0}&\textbf{32.1}&\textbf{45.7}\\
\cmidrule{1-1}\cmidrule(lr){2-3}\cmidrule{4-7}
\multirow{3}{*}{1.0}&\xmark&\xmark&17.6&27.0&17.3&21.8\\
&\cmark&\xmark&\underline{19.4}&\underline{40.8}&\underline{19.3}&\underline{27.3}\\
\rowcolor[HTML]{EFEFEF}{\cellcolor{white}{}} & \cellcolor{white}{\cmark} & 
\cellcolor{white}{\cmark} &\textbf{20.6}&\textbf{41.2}&\textbf{22.5}&\textbf{28.9}\\
\midrule
\end{tabular}
}}
\end{wraptable} 
\vspace{0.05in}
\noindent
\textbf{Adaptation Schemes. } 
In \cref{table:ablation_component}, we perform a component-wise analysis on our adaptation schemes (in \cref{sec:method:zero-shot-adaptation}): \textit{viz.}, 
fine-tuning (a) text-to-image diffusion model and (b) classifier. First, we confirm fine-tuning the text-to-image model obtains consistent gains in certified accuracy {across datasets} and perturbation bound $\varepsilon$, compared to the baseline without adapation.
The performance gains are further strengthened by also adapting the classifier.\footnote{We provide the complete results of certified accuracy across noise scale $\sigma$ and perturbation bound $\varepsilon$, in {\cref{table:ablation_other} of \cref{supple:abl-add}.}} 
We also remark that both adaptation schemes (\eqref{eq:dream-booth} and \eqref{eq:regularization-loss}) randomize the timestep $t\sim \mathcal{U}([0, T])$ during their fine-tuning, \ie, the observed gains in $\sigma=\{0.25, 0.5\}$ are from a single fine-tuning of the same pipeline but only using different timesteps at denoising. 

\label{sec:exp}
\section{Related Work}

\textbf{Adversarial Robustness. } Since the observation of adversarial examples \cite{szegedy2013intriguing,goodfellow2014explaining}, there have been continuous efforts in achieving adversarial robustness from neural networks, either in forms of \emph{empirical defenses} \cite{pmlr-v80-athalye18a,carlini2019evaluating,tramer2020adaptive} mainly based upon \emph{adversarial training} \cite{madry2018towards,zhang2019theoretically,Wang2020Improving,pmlr-v119-zhang20z,wu2020awp}, or \emph{certified defenses} \cite{wong2018provable,xiao2018training,cohen2019certified,zhang2020towards,leino2021globally}, depending on whether the robustness claim is provable or not. One of common beliefs in the literature has been that adversarial robustness is a property that has to be learned using concrete data \cite{schmidt2018adversarially,carmon2019unlabeled,gowal2021improving}. In this work, we move away from this assumption, proposing a novel direction of 
\jh{robustifying with no data}.

\vspace{0.05in}
\noindent
\textbf{Zero-shot Visual Recognition. }
Traditionally, zero-shot classification \cite{Palatucci2009zeroshot,Lampert2009zeroshot}, which aims to identify novel categories not present during training, has been a challenging task in computer vision. In recent years, large-scale vision-language models \cite{radford2021learning, jia2021scaling} have demonstrated remarkable capabilities in this regard, \eg, compared to prior arts \cite{Frome2013devise,Akata2015evalembed,Romera-Paredes2015EsZsl,ni2019dascn,Huang2019gdan,Schönfeld2019CADAVAE}. 
Upon these advances in obtaining high \emph{accuracy} under zero-shot, our work further questions whether it is possible to obtain \emph{adversarial robustness} in a zero-shot manner as well -- showing that large-scale text-to-image models can be a way to achieve this. 

\vspace{0.05in}
\noindent
\textbf{Transfer Learning for Robustness. }
Another line of research focuses on the \emph{transferability} of adversarial robustness, \ie, by utilizing an external source of data for robustness \cite{carmon2019unlabeled,gowal2021improving}. For example, several works \cite{dong2022fewshot, gold2020fewshot, wang2021fewshot, yin2018fewshot} considered a meta-learning based robust training aiming for \emph{few-shot} adaptation of adversarial robustness. Yet, they commonly rely on costly meta-training procedure from scratch, limiting their applicability to larger models.
More related to our work, Mao et al.~\citet{mao2023understanding} have recently proposed an adversarial contrastive fine-tuning for vision-language models, in order to transfer adversarial robustness to other zero-shot classification tasks. 
Again, in contrast to ours, the approach still requires a substantial amount of training data, \eg, as large as ImageNet in scale. 
\section{Conclusion}\label{sec:conclusion}
In this paper, we introduce a new formulation of robustifying vision classifiers {without external data}, as a more realistic concern in the era of adopting off-the-shelf models.
We propose a simple-yet-effective approach to this problem, which incorporates recent text-to-image diffusion models into the inference of a classifier in novel ways. Our approach is applicable for any off-the-shelf classifiers, making it as favorable and practical to obtain (provable) adversarial robustness when the use of external data is either limited or impractical for users. We hope our approach paves the way toward more reliable and secure AI systems, along the way towards robustifying existing components that have been considered powerful but fragile against adversarial attacks. 
We also believe our proposal suggests interesting future research, such as extending the framework to robustify commercial, black-box APIs \cite{salman2020denoised}. For example, this may require further techniques, such as zeroth-order optimization \cite{zhang2022zeroth}, as the adaptation scheme.

{\section*{Acknowledgements}
This work was partially supported by Center for Applied Research in Artificial Intelligence (CARAI) grant funded by Defense Acquisition Program Administration (DAPA) and Agency for Defense
Development (ADD) (UD230017TD), and by Institute for Information \& communications Technology Promotion (IITP) grant funded by the Korea government (MSIT) (No.RS-2019-II190075, Artificial Intelligence Graduate School Program (KAIST)). 
Jongheon Jeong acknowledges support from Institute of Information \& communications Technology Planning \& Evaluation (IITP) grant funded by the Korea government (MSIT) (No. RS-2019-II190079, Artificial Intelligence Graduate School Program (Korea University)), and from Korea University grant K2405671. 
We are grateful to the Center for AI
Safety (CAIS) for generously providing compute resources that supported a significant portion of the experiments conducted in this work. We thank Jihoon Tack for providing helpful feedbacks and suggestions in preparing an earlier version of our manuscript.}


%
\bibliographystyle{splncs04}
\bibliography{main}
\clearpage
\setcounter{page}{1}
\begin{center}
	\textbf{\large Adversarial Robustification via \\ Text-to-Image Diffusion Models \\}
    \vspace{0.1in}
    Appendix
\end{center}

\vspace{5pt}
\appendix
\section{Experimental Details}\label{supple:exp-detail}

\subsection{Datasets}
We basically consider a total of 12 datasets, including eight from the widely used zero-shot classification benchmarks \cite{radford2021learning, mao2023understanding}, {three from a more domain-specific benchmarks \cite{mohanty2016crop, helber2019_eurosat}} and one from ImageNet \cite{dataset/ilsvrc}. Specifically, these include: (a) STL \cite{coates2011stl10} and Caltech \cite{FeiFei2004caltech} for the classification of general objects, (b) Cars \cite{krause20133d}, Food \cite{bossard14food}, Pets \cite{Parkhi2012pets}, and Flower \cite{Nilsback2008flower} for domain-specific objects, and (c) SUN \cite{Xiao2010sun} for scene understanding, DTD \cite{dataset/dtd} as a textual benchmark inspired by human perception; as well as {(d) CropDiseases \cite{mohanty2016crop}, EuroSAT \cite{helber2019_eurosat} and ISIC \cite{codella2018isic} for a more specialized type of input.}

Following the approach of Mao et al. \cite{mao2023understanding}, we apply a $224\times224$ center cropping after rescaling to $256\times256$, except for STL, which is resized to $96\times96$. For evaluation, we subsample each dataset to approximately 500 samples, corresponding to the number of test samples for standard certification \cite{carlini2023diffusion, jeong2023multiscale}.

\subsection{Architectures}
We use DeepFloyd-IF\footnote{\url{https://github.com/deep-floyd/IF}} for the text-to-image diffusion model in our proposed framework. In particular, we adopt the IF-II-L checkpoint of DeepFloyd-IF as the super-resolution diffusion model. Throughout our experiments, we use the pre-trained CLIP-B/32 model and {an ImageNet pre-trained ResNet-50 model as the off-the-self classifiers to evaluate on.}

\subsection{Evaluation Metrics}
We evaluate adversarial robustness assuming $\ell_2$-adversary, considering two threat models of $\varepsilon \in \{0.5, 1.0\}$. When comparing with Mao et al.~\citet{mao2023understanding}, the major empirical defense baseline we consider, we report the \emph{empirical robust accuracy} from 100-step projected gradient descent (PGD-100) \cite{madry2018towards}, following Mao et al.~\citet{mao2023understanding}. We use the step size of $\alpha=\frac{4}{3}\cdot\frac{\varepsilon}{\textit{\# steps}}$ here, and adopt the other attack hyperparameters from Mao et al.~\citet{mao2023understanding}. To measure the empirical robust accuracy of smoothed classifiers, including our method, we apply SmoothAdv \cite{salman2019smoothadv} with $m_{test} = 32$ Gaussian noise samples as an adaptive attack scheme at PGD instead of directly attacking the base classifier, in attempt to maximize the success rate of attacks \cite{tramer2020adaptive}. {For PGD against the denoise-and-classify pipeline our method uses, we consider the full gradient propagation including the denoising process. This is computationally feasible because our method is based on \emph{single-step}
denoising as described in Eq. (9).}
When making predictions from smoothed classifiers, we follow the standard protocol of Cohen et al.~\citet{cohen2019certified}. More concretely, they proposed a Monte Carlo-based procedure, namely \textsc{Predict}, which estimates the prediction using $n$ noise samples and outputs only when it is statistically consistent with the true $\hat{f}(x)$, given the randomness of $n$ samples and significance level $\alpha$: otherwise, it ``{abstains}'' from making a prediction. We use $n=100$ and $\alpha = 0.001$ for $\textsc{Predict}$. When a smoothed classifier makes abstention for a test sample, we re-evaluate the sample using the base classifier $f$ in measuring empirical clean accuracy, following the protocol of Jeong et al.~\citet{jeong2023multiscale}. Hence, we report the \textit{empirical clean accuracy} as the fraction of test samples that are either (a) correct with $\textsc{Predict}$, or (b) abstained but still correct at the base classifier $f$.

In addition to empirical accuracy, we also report \emph{certified robust accuracy} \cite{cohen2019certified} when reporting smoothed classifiers, to show the ``lower-bound'' of robust accuracy that the classifier achieves for \emph{every possible} empirical attack to $\hat{f}$. In this way, one can rule out the possibility that a stronger attack over PGD-100 we consider, \eg AutoAttack \cite{croce2020auto}, might refute the robustness claims on $\hat{f}$. Similarly to $\textsc{Predict}$, we adopt $\textsc{Certify}$ \cite{cohen2019certified} for this procedure, using $n_0 = 100$, $n = 10, 000$ and $\alpha = 0.001$. More details on the certification of smoothed classifiers are given in \cref{supple:pred-cer}.

\subsection{Implementation}
\begin{figure}[t]
\centering
\begin{subfigure}{0.475\columnwidth}
\includegraphics[width=\linewidth]{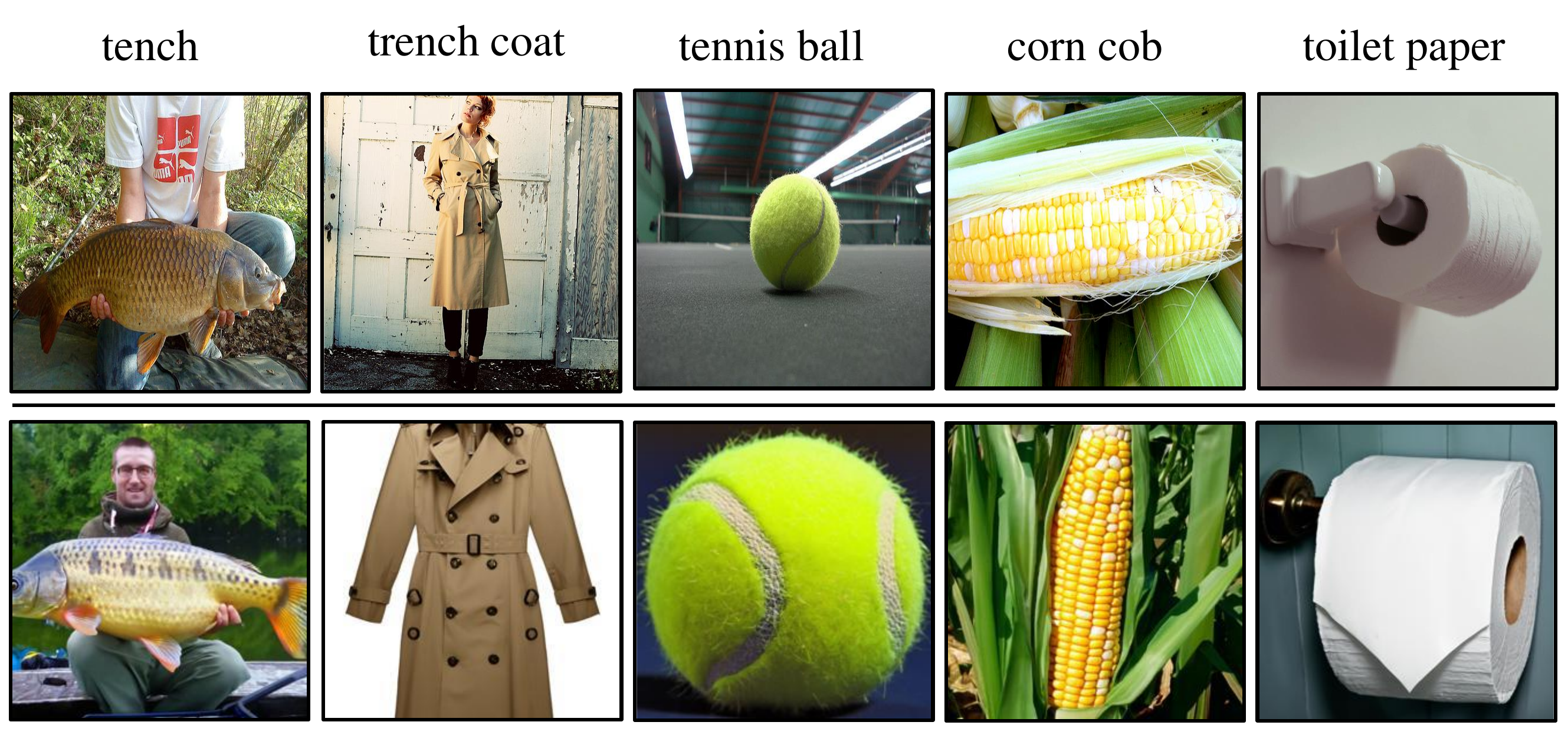}
\caption{ImageNet \cite{dataset/ilsvrc}}
\end{subfigure}
\begin{subfigure}{0.475\columnwidth}
\includegraphics[width=\linewidth]{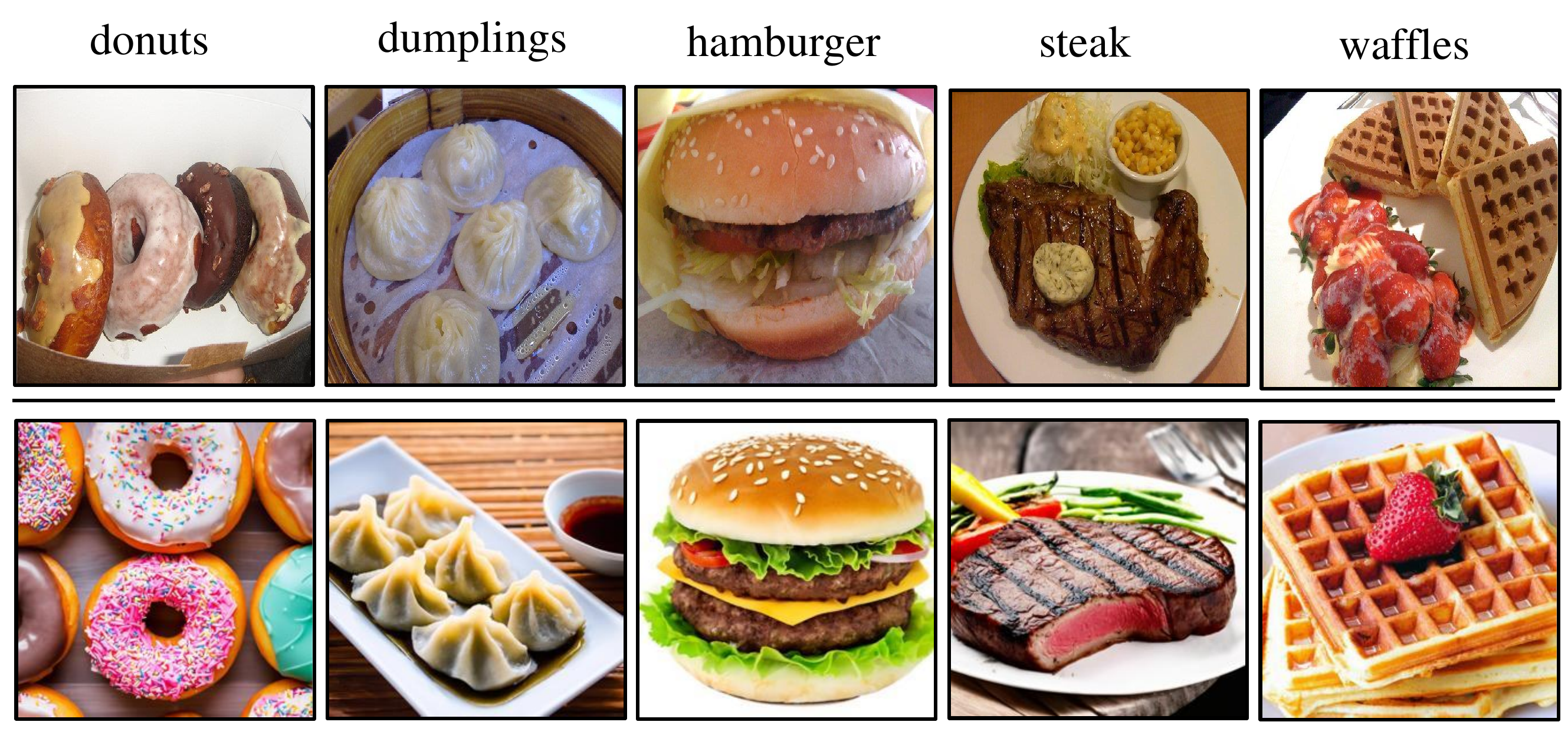}
\caption{Food \cite{bossard14food}}
\end{subfigure}
\begin{subfigure}{0.475\columnwidth}
\includegraphics[width=\linewidth]{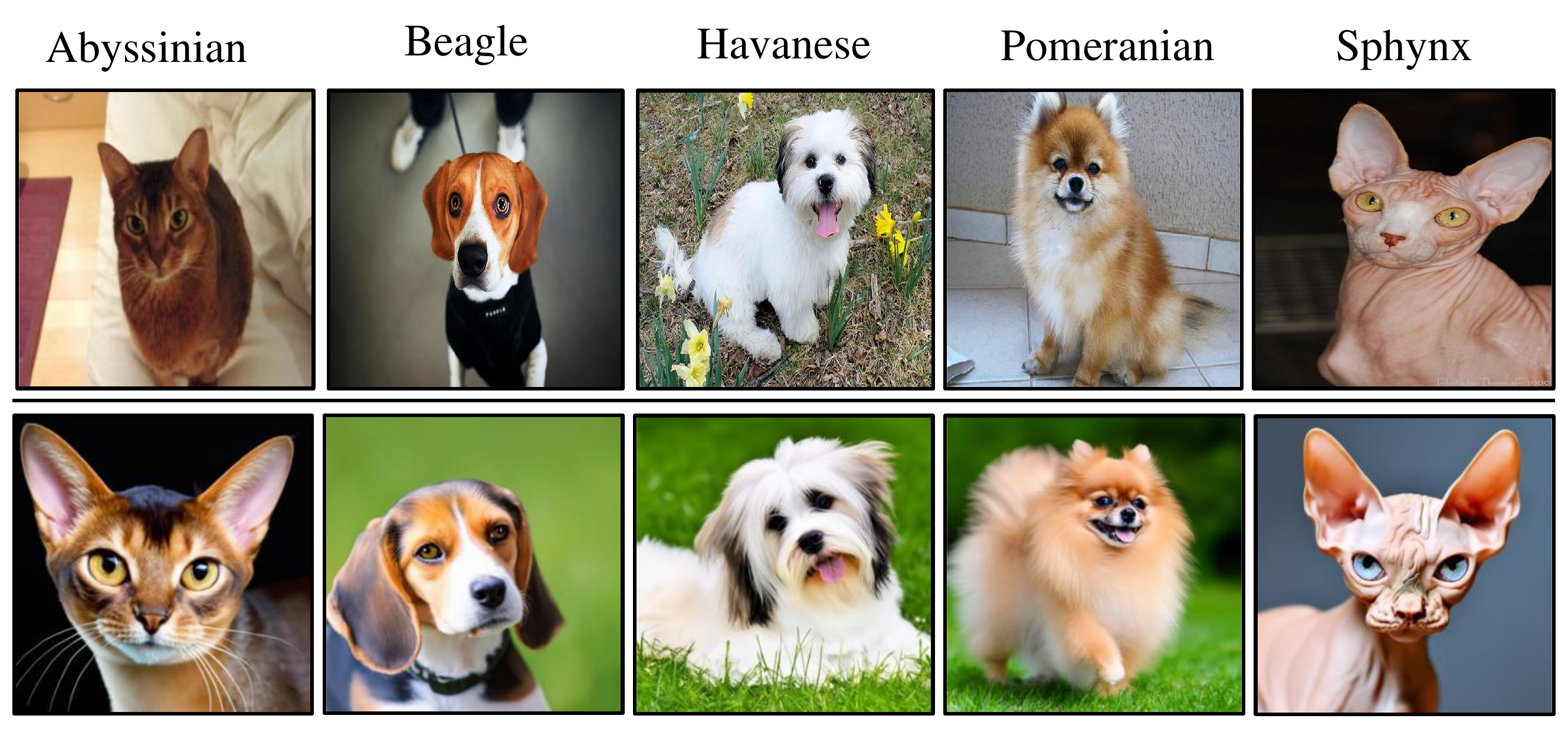}
\caption{Pet \cite{Parkhi2012pets}}
\end{subfigure}
\begin{subfigure}{0.475\columnwidth}
\includegraphics[width=\linewidth]{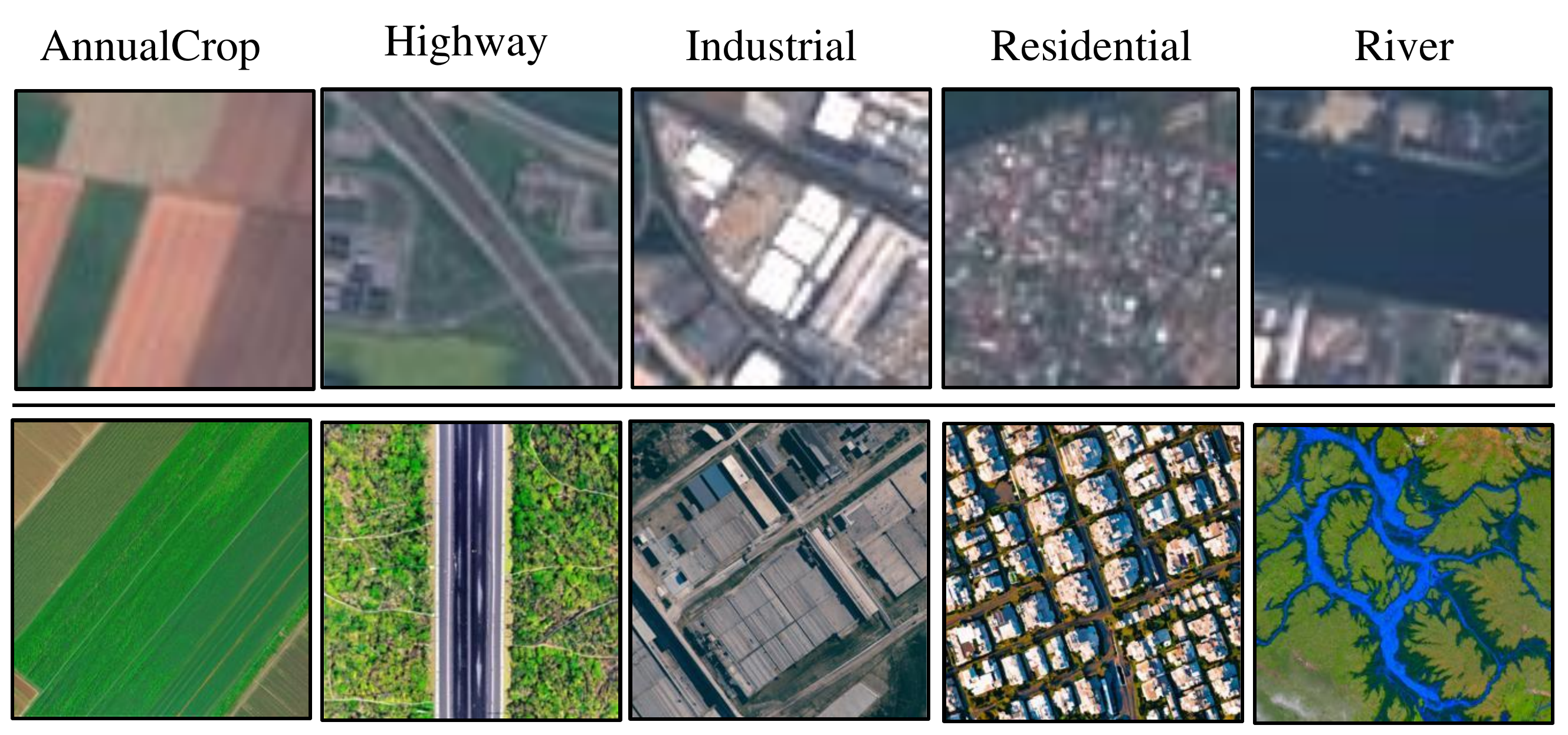}
\caption{EuroSAT \cite{helber2019_eurosat}}
\end{subfigure}
\caption{Qualitative comparison between original images and synthetic images across various datasets. The first row displays the original images from the each dataset and the second row presents the corresponding synthetic images.}
\vspace{-0.2in}
\label{figure:synthetic-images}
\end{figure}

\textbf{Reference Set Synthesis.} We use checkpoints from DeepFloyd-IF to synthesize the reference set. Specifically, we employ checkpoint IF-I-XL for the low-resolution diffusion model and IF-II-L for the super-resolution diffusion model. Subsequently, we generate $256\times256$ images by sequentially passing through these checkpoints. {We synthesize one
image per class for adaptation, \ie, we consider 1-shot adaptation by default, although it is possible to apply our scheme for higher-shot setups (\eg, as explored in \cref{table:ablation-reference}).} Examples of the synthetic reference set are provided in \cref{figure:synthetic-images}.

\vspace{0.05in}
\noindent
\textbf{Diffusion Personalization. }
We follow the \textit{DreamBooth} implementation from the \texttt{diffusers} library released by Huggingface, which is available at \url{https://huggingface.co/docs/diffusers/training/dreambooth}. Specifically, we fine-tune the DeepFloyd-IF checkpoint IF-II-L for 500 training steps, using a learning rate of $1\times10^{-4}$ and a batch size of 24. For \textit{classifier-guided} regularization, we use $\lambda=0.01$.

\vspace{0.05in}
\noindent
\textbf{Classifier Fine-tuning. }
We follow Mao et al.~\citet{mao2023understanding} to fine-tune CLIP-B/32. In contrast to its protocol, which involves solely fine-tuning the image-encoder of CLIP, we fine-tune both the text and image-encoder over 10 training epochs, using a learning rate of $5\times10^{-7}$ and a batch size of 256.

\vspace{0.05in}
\noindent
\textbf{Denoised Smoothing. }
To apply randomized smoothing, an input $x\in [0, 1]^{c\cdot h\cdot w}$ is given as corrupted $\hat{x}:= x + \delta$ with noise $\delta\sim\mathcal{N}(0,\sigma^{2}\textbf{I})$. For denoised smoothing, $x$ is first normalized to $[-1, 1]^{c\cdot h\cdot w}$ with the mean $[0.5, 0.5, 0.5]$ and standard deviation $[0.5, 0.5, 0.5]$, the standard training configurations of diffusion models. Then, the \textit{timestep} $\hat{t}$ is estimated by \eqref{eq:relation-alpha} to correspond $\hat{x}$ to $x_{\hat{t}}$. When computing \eqref{eq:relation-alpha}, it is necessary to multiply the noise strength $\sigma$ by 2. This adjustment is made because the noise $\delta$ associated with $x \in [0, 1]^{c \cdot h \cdot w}$ is doubled due to the normalization.

\vspace{0.05in}
\noindent
{\textbf{$\ell_2$-based Models of Mao et al.~\cite{mao2023understanding}.} The official code released by Mao et al.~\cite{mao2023understanding} implements fine-tuning with an $\ell_2$-adversary, and we follow the code whenever reporting results for Mao et al.~\cite{mao2023understanding}.}

\subsection{Prediction and Certification}\label{supple:pred-cer}
\begin{figure}[t]\centering
\setlength{\tabcolsep}{5pt}
{\small
\begin{minipage}[t]{0.75\columnwidth}\centering
\begin{algorithm}[H]
\caption{\textsc{Predict} and \textsc{Certify}}
\label{algro:predict-certify}
\begin{algorithmic}[1]

\Function{\textsc{Predict}}{$f, \sigma, x, n, \alpha$}
\State $\mathtt{counts} \leftarrow \textsc{SampleUnderNoise}(f, x, n, \sigma)$
\State $\hat{c}_A, \hat{c}_B \leftarrow \text{top two indices in $\mathtt{counts}$}$
\State $n_A, n_B \leftarrow \mathtt{counts}[\hat{c}_A], \mathtt{counts}[\hat{c}_B]$
\If{$\textsc{BinomPValue}(n_A, n_A + n_B, 0.5)\leq\alpha$}
\State\Return$\hat{c}_A$
\Else
\State\Return\textsc{ABSTAIN}
\EndIf 
\EndFunction

\Function{\textsc{Certify}}{$f, \sigma, x, n_0, n, \alpha$}
\State $\mathtt{counts0} \leftarrow \textsc{SampleUnderNoise}(f, x, n_0, \sigma)$
\State $\hat{c}_A \leftarrow \text{top index in $\mathtt{counts0}$}$
\State $\mathtt{counts} \leftarrow \textsc{SampleUnderNoise}(f, x, n, \sigma)$
\State \hbox{$\underline{p_{A}}
\leftarrow \textsc{LowConfBound}(\mathtt{counts}[\hat{c}_A], n, 1-\alpha)$}

\If{$\underline{p_{A}}>\frac{1}{2}$}
\State\Return prediction $\hat{c}_A$ and radius $\sigma\cdot \Phi^{-1}(\underline{p_{A}})$
\Else
\State\Return\textsc{ABSTAIN}
\EndIf 
\EndFunction
\end{algorithmic}
\end{algorithm}
\end{minipage}
}
\vspace{-0.2in}
\end{figure}

\vspace{0.05in}
\noindent
Given a classifier $f$, prediction and certification of the smoothed classifier $\hat{f}$ are approximated using practical Monte Carlo algorithms, following Cohen et al. \citet{cohen2019certified}. These procedures are provided as \textsc{Predict} and \textsc{Certify} in Algorithm~\ref{algro:predict-certify}. Here, the $\textsc{SampleUnderNoise}(f, x, n, \sigma)$ function returns an array where each element represents the count of predictions made by $f$ on input $x$ under each of the $n$ trials of noise sampling from $\mathcal{N}(0, \sigma^{2}\textbf{I})$. In \textsc{Predict}, the smoothed classifier $\hat{f}$ returns class $\hat{c}_A$ if $\hat{c}_A$ is predicted more often than other classes in $n$ trials. The criterion of ``more often'' is decided by whether $\textsc{BinomPValue}(n_A, n_A + n_B, p)$, returning the p-value of the two-sided hypothesis test that $n_A\sim\text{Binomial}(n_A + n_B, p)$, is less than or equal to the threshold $\alpha$. \textsc{Predict} is regarded as inference of $\hat{f}$ when we practically use it. On the other hand, we use \textsc{Certify} when we want to not only make predictions but also compute the robustness (the radius in Algorithm~\ref{algro:predict-certify}) of $\hat{f}$ on input $x$. This process involves the estimation of the lower bound $\underline{p_A}$ on the probability that $f$ predicts $\hat{c}_A$ under noise. It is computed by $\textsc{LowConfBound}(n_A, n, 1-\alpha)$, which returns a one-sided $(1-\alpha)$ lower confidence interval for the parameter $p$ given a sample $n_A\sim\mathrm{Binomial}(n, p)$. In the context of \textit{denoised smoothing} \cite{salman2020denoised}, the classifier $f$ can be a pipeline that combines a denoiser with a standard classifier.

\subsection{Computational Resources}
We conduct experiments with a cluster consisting of 4 NVIDIA A100 80GB GPUs. The synthesis of the reference set is executed on a single NVIDIA A100 80GB GPU, which typically requires $\sim2$ minutes per image. In applying our self-adaptation schemes and inference, we use 4 NVIDIA A100 80GB GPUs. The execution time of the adaptation schemes can be influenced by the size of the reference set, proportional to the number of classes. For ImageNet, having the largest reference set in experiments, a single run takes $\sim20$ minutes for the diffusion personalization and $\sim30$ minutes for the CLIP fine-tuning. For a single run of the certification process in our framework, we observe $\sim4$ minutes of per-image cost with $n = 10,000$.

\begin{figure}[t]\centering
\setlength{\tabcolsep}{5pt}
{\small
\begin{minipage}[t]{\columnwidth}\centering
\begin{algorithm}[H]
\caption{Adversarial Robustification via Text-to-Image Diffusion Models }
\label{algro:zero-ar}
\begin{algorithmic}[1]

\Require Textual label set $\mathcal{Y}=\{c_i\}_{i=1}^K$, textual template $\mathtt{C}()$, correction factor $k$, weight hyperparameter $\lambda$, learning rates $\alpha$, $\beta$.

\algrule

\For{$i=1$ to $K$}
\State Generate a image $x_i^g$ using $\mathtt{C}(c_i)$ into the text-to-image diffusion model. 
\EndFor

\State Construct synthetic reference set $D^g = \{(x_i^{g}, c_i)\}_{i=1}^K$.

\While{not done}
\State Sample mini-batch $\mathcal{B}=\{(x^g_b, c_b )\}_{b=1}^{B}$ from $D^g$
\For{$b=1$ to $B$}
\State Sample a timestep $t\sim\mathcal{U}([0, T])$
\State Sample a Gaussian noise $\varepsilon\sim\mathcal{N}(0,\textbf{I})$
\State $x^g_{t, b} = \sqrt{\alpha_t}\cdot x^g_b + \sqrt{1-\alpha_t}\cdot\varepsilon$
\Comment{\eqref{eq:forward pass}}

\State $\hat{\varepsilon} = \varepsilon_{\theta}(x^g_{t, b}, t, \tau_{\theta}(\mathtt{C}(``sks"))|x^g_{t, b}, kt)$

\State $\tilde{x}^g_{b} = \frac{1}{\sqrt{\alpha_t}}\cdot x^g_{t, b} - \sigma\cdot \hat{\varepsilon}$  
\Comment{Eq. (\ref{eq:one-step-denoising-super})}

\State $L_{\tt diff}^{(b)}(\theta)=||\varepsilon - \hat{\varepsilon}||^2_2$
\Comment{\eqref{eq:dream-booth}}
\State $L_{\tt clf}^{(b)}(\theta, \psi) = \mathbb{CE}(f_{\psi}(\tilde{x}^g_b), c_b)$
\Comment{\eqref{eq:classifier-guided loss}}
\EndFor
\State $\theta \leftarrow \theta - \frac{\alpha}{B} \sum_{b=1}^B\{{L}_{\tt diff}^{(b)}(\theta) + \lambda \cdot L_{\tt clf}^{(b)}(\theta, \psi)\}$
\\
\Comment{Update the denoiser network
$\varepsilon_{\theta}$}
\State $\psi \leftarrow \psi - \frac{\beta}{B} \sum_{b=1}^B L_{\tt clf}^{(b)}(\theta, \psi)$
\\
\Comment{Update the classifier
$f_{\psi}$}

\EndWhile

\end{algorithmic}
\end{algorithm}
\end{minipage}
}
\vspace{-0.2in}
\end{figure}

\clearpage

\section{Overall Procedure}
Given a textual label set $\mathcal{Y}$, we first fine-tune the denoiser network $\varepsilon_{\theta}$ of the text-to-image super-resolution diffusion model and the classifier $f_{\psi}$. This process is outlined in Algorithm~\ref{algro:zero-ar}. After fine-tuning, we can perform predictions on input $x$ using denoised smoothing, which consists of the optimized denoiser network $\varepsilon_{\theta^*}$ and the classifier $f_{\psi^*}$. Through this overall framework, we ensure that any off-the-shelf classifiers achieve strong and provable adversarial robustness on input $x$ using only a textual label set.

\begin{table}
\centering
\vspace{0.2in}
\caption{
Ablation study on adaptation schemes in other datasets. Bold indicates the best and runner-up is underlined.
}\label{table:ablation_other}
\vspace{-0.2in}
\begin{subtable}{0.475\columnwidth}
\centering
\setlength{\tabcolsep}{5pt}
\caption{ImageNet \cite{dataset/ilsvrc}}
\vspace{-0.1in}
\scalebox{0.55}{
\begin{tabular}{cccccccccc}
\toprule
\multirow{2}{*}{}& \multicolumn{2}{c}{Adaptation} & & \multicolumn{6}{c}{Certified accuracy at $\varepsilon$ (\%)}\\
\cmidrule(lr){2-3} \cmidrule{5-10}
$\sigma$& T2I & CLIP &ACR&0.0&0.25&0.5&0.75&1.0&1.25\\
\cmidrule{1-1}\cmidrule(lr){2-3}\cmidrule(lr){4-4}\cmidrule{5-10}
\multirow{3}{*}{0.25}&\xmark&\xmark&0.277&50.8&41.8&29.6&19.2\\
&\cmark&\xmark&0.292&\underline{52.2}&\underline{43.0}&\underline{31.8}&\underline{21.4}\\
\rowcolor[HTML]{EFEFEF}{\cellcolor{white}{}} & \cellcolor{white}{\cmark} & \cellcolor{white}{\cmark} &\textbf{0.303}&\textbf{53.4}&\textbf{44.8}&\textbf{34.2}&\textbf{21.6}&&\\
\cmidrule{1-1}\cmidrule(lr){2-3}\cmidrule(lr){4-4}\cmidrule{5-10}
\multirow{3}{*}{0.50}&\xmark&\xmark&0.367&42.0&35.4&29.2&23.8&17.6&13.0\\
&\cmark&\xmark&0.394&\underline{44.0}&\underline{37.0}&\underline{30.8}&\textbf{25.2}&\underline{19.4}&\underline{15.2}\\
\rowcolor[HTML]{EFEFEF}{\cellcolor{white}{}} & \cellcolor{white}{\cmark} & 
\cellcolor{white}{\cmark} &\textbf{0.409}&\textbf{46.0}&\textbf{38.8}&\textbf{32.4}&\underline{25.0}&\textbf{20.6}&\textbf{15.4}\\
\midrule
\end{tabular}}
\vspace{0.05in}
\end{subtable}
\begin{subtable}{0.475\columnwidth}
\centering
\setlength{\tabcolsep}{5pt}
\caption{STL \cite{coates2011stl10}}
\vspace{-0.1in}
\scalebox{0.55}{
\begin{tabular}{cccccccccc}
\toprule
\multirow{2}{*}{}& \multicolumn{2}{c}{Adaptation} & & \multicolumn{6}{c}{Certified accuracy at $\varepsilon$ (\%)}\\
\cmidrule(lr){2-3} \cmidrule{5-10}
$\sigma$& T2I & CLIP &ACR&0.0&0.25&0.5&0.75&1.0&1.25\\
\cmidrule{1-1}\cmidrule(lr){2-3}\cmidrule(lr){4-4}\cmidrule{5-10}
\multirow{3}{*}{0.25}&\xmark&\xmark&0.502&86.8&74.2&55.2&37.2\\
&\cmark&\xmark&0.568&\underline{90.0}&\underline{80.4}&\bf{66.0}&\bf{45.6}\\
\rowcolor[HTML]{EFEFEF}{\cellcolor{white}{}} & \cellcolor{white}{\cmark} & \cellcolor{white}{\cmark} &\textbf{0.570}&\textbf{90.8}&\textbf{80.8}&\textbf{66.0}&\underline{45.0}&&\\
\cmidrule{1-1}\cmidrule(lr){2-3}\cmidrule(lr){4-4}\cmidrule{5-10}
\multirow{3}{*}{0.50}&\xmark&\xmark&0.579&68.0&60.0&47.4&38.4&27.0&19.2\\
&\cmark&\xmark&0.783&\underline{79.8}&\bf{72.0}&\underline{61.0}&\textbf{52.2}&\underline{40.8}&\underline{31.4}\\
\rowcolor[HTML]{EFEFEF}{\cellcolor{white}{}} & \cellcolor{white}{\cmark} & 
\cellcolor{white}{\cmark} &\textbf{0.787}&\textbf{80.2}&\underline{71.6}&\textbf{61.8}&\underline{51.8}&\textbf{41.2}&\textbf{32.0}\\
\midrule
\end{tabular}}
\vspace{0.05in}
\end{subtable}
\begin{subtable}{0.475\columnwidth}
\centering
\setlength{\tabcolsep}{5pt}
\caption{SUN \cite{Xiao2010sun}}
\vspace{-0.1in}
\scalebox{0.55}{
\begin{tabular}{cccccccccc}
\toprule
\multirow{2}{*}{}& \multicolumn{2}{c}{Adaptation} & & \multicolumn{6}{c}{Certified accuracy at $\varepsilon$ (\%)}\\
\cmidrule(lr){2-3} \cmidrule{5-10}
$\sigma$& T2I & CLIP &ACR&0.0&0.25&0.5&0.75&1.0&1.25\\
\cmidrule{1-1}\cmidrule(lr){2-3}\cmidrule(lr){4-4}\cmidrule{5-10}
\multirow{3}{*}{0.25}&\xmark&\xmark&0.260&50.3&38.0&28.3&16.1\\
&\cmark&\xmark&0.277&\underline{51.8}&\underline{41.4}&\underline{30.7}&\underline{17.5}\\
\rowcolor[HTML]{EFEFEF}{\cellcolor{white}{}} & \cellcolor{white}{\cmark} & \cellcolor{white}{\cmark} &\textbf{0.293}&\textbf{55.4}&\textbf{44.0}&\textbf{32.1}&\textbf{18.9}&&\\
\cmidrule{1-1}\cmidrule(lr){2-3}\cmidrule(lr){4-4}\cmidrule{5-10}
\multirow{3}{*}{0.50}&\xmark&\xmark&0.381&46.0&39.2&31.9&24.3&17.3&10.8\\
&\cmark&\xmark&0.420&\underline{48.4}&\underline{42.6}&\underline{36.7}&\underline{26.9}&\underline{19.3}&\underline{12.0}\\
\rowcolor[HTML]{EFEFEF}{\cellcolor{white}{}} & \cellcolor{white}{\cmark} & 
\cellcolor{white}{\cmark} &\textbf{0.453}&\bf{53.6}&\textbf{46.0}&\textbf{37.8}&\bf{29.0}&\textbf{22.5}&\textbf{12.9}\\
\midrule
\end{tabular}}
\vspace{0.05in}
\end{subtable}
\begin{subtable}{0.475\columnwidth}
\centering
\setlength{\tabcolsep}{5pt}
\caption{Food \cite{bossard14food}}
\vspace{-0.1in}
\scalebox{0.55}{
\begin{tabular}{cccccccccc}
\toprule
\multirow{2}{*}{}& \multicolumn{2}{c}{Adaptation} & & \multicolumn{6}{c}{Certified accuracy at $\varepsilon$ (\%)}\\
\cmidrule(lr){2-3} \cmidrule{5-10}
$\sigma$& T2I & CLIP &ACR&0.0&0.25&0.5&0.75&1.0&1.25\\
\cmidrule{1-1}\cmidrule(lr){2-3}\cmidrule(lr){4-4}\cmidrule{5-10}
\multirow{3}{*}{0.25}&\xmark&\xmark&0.382&71.2&55.8&43.6&22.8\\
&\cmark&\xmark&0.411&\bf{74.5}&\underline{61.2}&\underline{43.8}&\underline{26.9}\\
\rowcolor[HTML]{EFEFEF}{\cellcolor{white}{}} & \cellcolor{white}{\cmark} & \cellcolor{white}{\cmark} &\textbf{0.416}&\textbf{74.5}&\textbf{61.8}&\textbf{45.7}&\textbf{27.3}&&\\
\cmidrule{1-1}\cmidrule(lr){2-3}\cmidrule(lr){4-4}\cmidrule{5-10}
\multirow{3}{*}{0.50}&\xmark&\xmark&0.479&57.4&47.1&39.8&30.9&21.8&16.0\\
&\cmark&\xmark&0.555&\bf{65.0}&\underline{54.9}&\underline{44.4}&\underline{36.0}&\underline{27.3}&\underline{18.0}\\
\rowcolor[HTML]{EFEFEF}{\cellcolor{white}{}} & \cellcolor{white}{\cmark} & 
\cellcolor{white}{\cmark} &\textbf{0.563}&\underline{64.8}&\textbf{55.4}&\textbf{45.1}&\bf{36.4}&\textbf{28.9}&\textbf{19.0}\\
\midrule
\end{tabular}}
\end{subtable}
\vspace{-0.1in}
\end{table}

\section{Additional Ablation Study}\label{supple:abl-add}

\noindent
\textbf{Adaptation Schemes. }
In \cref{table:ablation_component}, we have validated the effectiveness of our self-adaptation schemes {across multiple datasets under $\ell_2$-adversary of $\varepsilon=0.5$ and $1.0$}. These schemes involve fine-tuning both the (a) text-to-image diffusion model and (b) classifier.
In \cref{table:ablation_other}, we provide {the full results of them}, demonstrating that these adaptations significantly improve baseline (not using adaptation) performance across $\sigma$ and $\varepsilon$. The results highlight again the versatility of our adaptation schemes, demonstrating their effectiveness on diverse tasks.

\begin{table}[t]
\centering
\setlength{\tabcolsep}{5pt}
\hspace*{\fill}
\parbox[t]{0.475\textwidth}{
\centering\small
\caption{
Ablation study on $\lambda$ in \eqref{eq:final-person-loss}. Bold indicates the best.}\label{table:ablation-lambda}
\vspace{-0.05in}
\resizebox{0.45\textwidth}{!}{
\begin{tabular}{ccccccccc}
\toprule
\multirow{2}{*}{}&&&\multicolumn{6}{c}{Certified accuracy at $\varepsilon$ (\%)}\\
\cmidrule{4-9}
$\sigma$&$\lambda$&ACR&0.0&0.25&0.5&0.75&1.0&1.25\\
\cmidrule{1-1}\cmidrule(lr){2-2}\cmidrule(lr){3-3}\cmidrule{4-9}
&0.0\phantom{0}\phantom{0}&0.270&49.6&39.0&30.0&19.6\\
0.25 &0.001&0.280&50.6&40.2&30.8&20.2&&\\
\rowcolor[HTML]{EFEFEF}{\cellcolor{white}{}} & \cellcolor{white}{0.01\phantom{0}}  &\textbf{0.292}&\textbf{52.2}&\textbf{43.0}&\textbf{31.8}&\textbf{21.4}&&\\
&0.1\phantom{0}\phantom{0}&0.290&51.8&42.8&31.2&20.6\\
\midrule
&0.0\phantom{0}\phantom{0}&0.358&38.0&32.6&27.8&22.2&18.0&14.4\\
0.50 &0.001&0.379&40.4&35.0&30.0&24.6&\textbf{20.0}&14.6\\
\rowcolor[HTML]{EFEFEF}{\cellcolor{white}{}} & \cellcolor{white}{0.01\phantom{0}}  &\textbf{0.394}&\textbf{44.0}&\textbf{37.0}&\textbf{30.8}&\textbf{25.2}&19.4&15.2\\
&0.1\phantom{0}\phantom{0}&0.390&43.4&37.0&30.4&24.2&\textbf{20.0}&\textbf{15.4}\\
\bottomrule
\end{tabular}}}
~~~~
\parbox[t]{0.45\textwidth}{
\centering
\setlength{\tabcolsep}{5pt}
\small
\caption{
Ablation study on size of reference set. Here,  \textit{shot} means that how many generated instances per class in synthetic reference set. 
The bold indicates the best.}\label{table:ablation-reference}
\vspace{-0.05in}
\resizebox{0.4\textwidth}{!}{
\begin{tabular}{ccccccccc}
\toprule
\multirow{2}{*}{}&&&\multicolumn{6}{c}{Certified accuracy at $\varepsilon$ (\%)}\\
\cmidrule{4-9}
$\sigma$&\textit{shot}&ACR&0.0&0.25&0.5&0.75&1.0&1.25\\
\cmidrule{1-1}\cmidrule(lr){2-2}\cmidrule(lr){3-3}\cmidrule{4-9}
\multirow{2}{*}{0.25}&1 &0.303&53.4&\textbf{44.8}&34.2&21.6\\
&4&\cellcolor[HTML]{EFEFEF}{\textbf{0.309}}&\cellcolor[HTML]{EFEFEF}{\textbf{54.2}}&\cellcolor[HTML]{EFEFEF}{44.2}&\cellcolor[HTML]{EFEFEF}{\textbf{35.4}}&\cellcolor[HTML]{EFEFEF}{\textbf{24.8}}&\cellcolor[HTML]{EFEFEF}{}&\cellcolor[HTML]{EFEFEF}{}\\
\midrule
\multirow{2}{*}{0.5}&1&0.409&\textbf{46.0}&38.8&32.4&25.0&20.5&15.4\\
&4&\cellcolor[HTML]{EFEFEF}{\textbf{0.428}}&\cellcolor[HTML]{EFEFEF}{45.2}&\cellcolor[HTML]{EFEFEF}{\textbf{40.2}}&\cellcolor[HTML]{EFEFEF}{\textbf{34.8}}&\cellcolor[HTML]{EFEFEF}{\textbf{27.2}}&\cellcolor[HTML]{EFEFEF}{\textbf{21.6}}&\cellcolor[HTML]{EFEFEF}{\textbf{17.0}}\\
\bottomrule
\end{tabular}}}
\hspace*{\fill}
\end{table}

\vspace{0.05in}
\noindent
\textbf{Classifier-Guided Regularization.} To test the effectiveness of our proposed regularization loss in \eqref{eq:regularization-loss}, we ablate the regularization strength $\lambda$ from $0.0$ to $0.1$ in fine-tuning text-to-image diffusion model, where $\lambda = 0.01$ is used by default in our experiments. As shown in \cref{table:ablation-lambda}, we observe using higher values of $\lambda$, \eg, $\lambda=0.01$ or $0.1$ achieve higher ACR as well as overall certified accuracy compared to $\lambda=0.0$, confirming the effect of the proposed regularization in our self-personalization procedure.

\vspace{0.05in}
\noindent
\textbf{Synthetic References. } 
To evaluate the impact of reference set size on our framework, we conduct the experiment on ImageNet using a larger set, generating four images per class. In \cref{table:ablation-reference}, we observe that utilizing a larger reference set size (4-shot) outperforms our default size (1-shot). This finding highlights the scalability of our framework, indicating that additional performance gains can be achieved as the size of the reference set increases.

\vspace{0.05in}
\noindent
\subsection{Analysis of Correction Factor}\label{supple:correct-factor}
In \cref{figure:plot_correct_factor}, we ablate the effect of correction factor $k$, varying in the range $0.0$ to $2.4$ with a step size of $0.2$. We focus on comparing the accuracy of denoised images from a subset of ImageNet. Specifically, each sample is perturbed with Gaussian noise of $\sigma=0.25$ and $0.5$ and denoised through our zero-shot denoising \eqref{eq:one-step-denoising-super}. 
In \cref{figure:plot_correct_factor}, we observe that higher correction factor yields greater accuracy compared to lower correction factor ($< 1$). Further, we decide to adopt a correction factor $k$ of $1.8$ within our framework.

\begin{figure}[t]
\vspace{0.15in}
\centering
\begin{subfigure}{0.475\columnwidth}
\includegraphics[width=\linewidth]{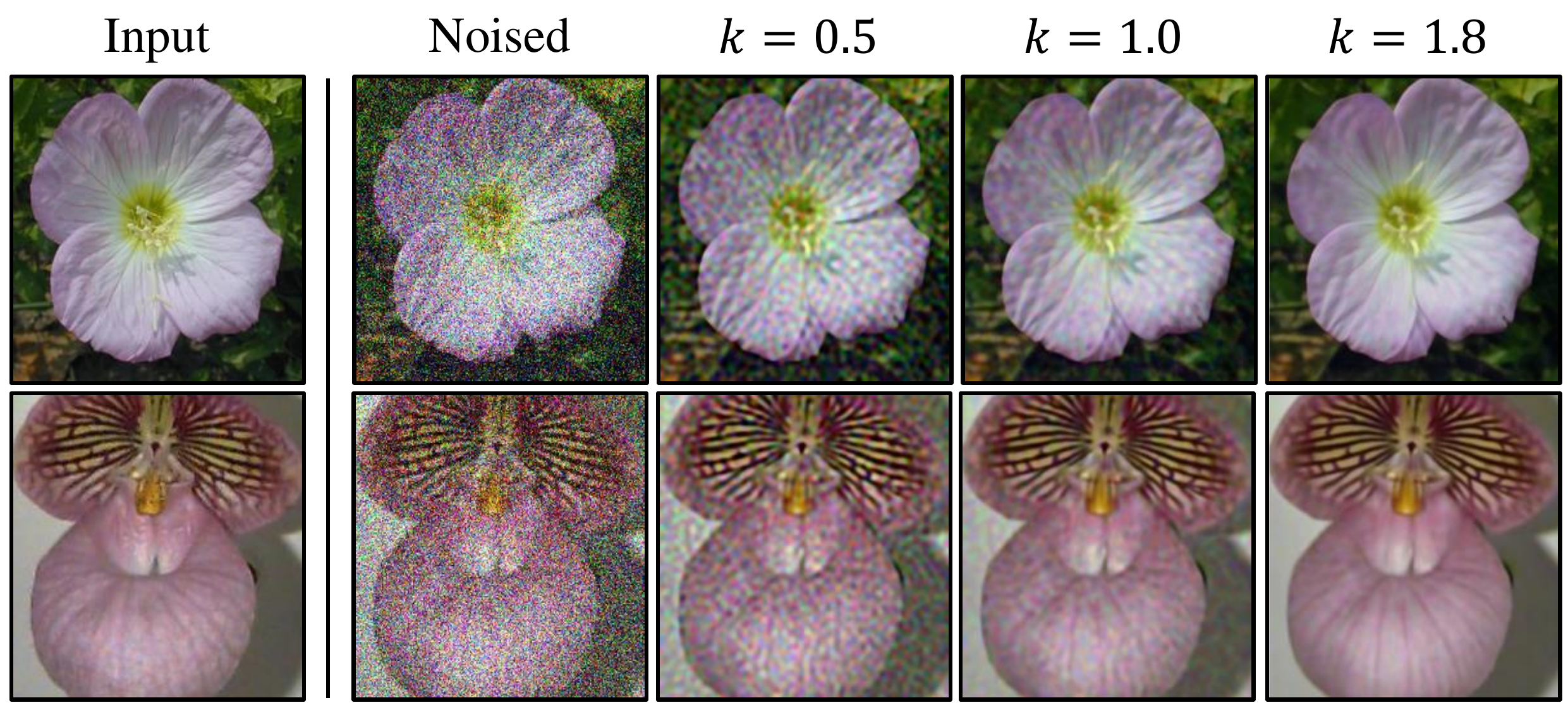}
\caption{Flower \cite{Nilsback2008flower}}

\end{subfigure}
\begin{subfigure}{0.475\columnwidth}
\includegraphics[width=\linewidth]{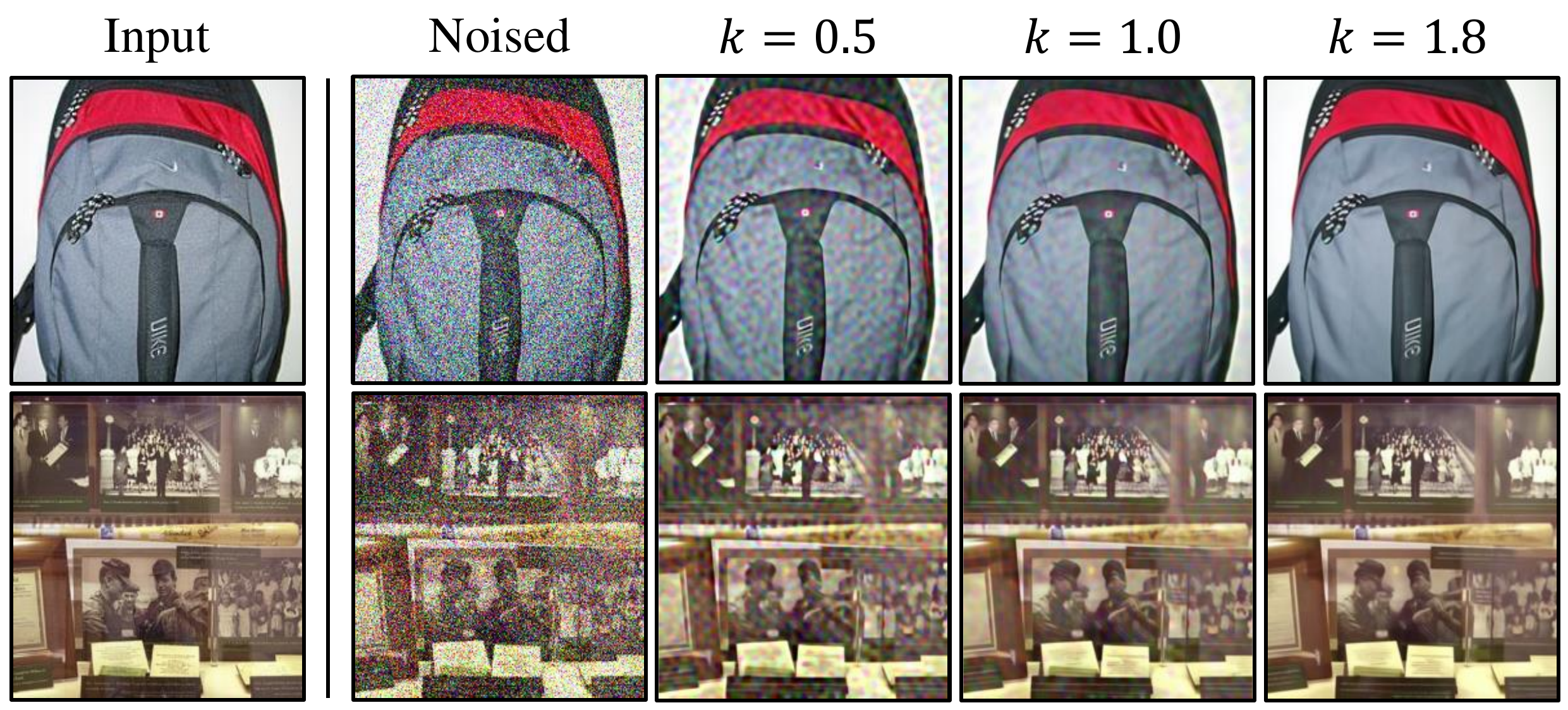}
\caption{Caltech \cite{FeiFei2004caltech}}
\end{subfigure}
\caption{Qualitative comparisons of denoised images on varying correct factor $k$ in other datasets. We perturb each input using Gaussian noise of $\sigma=0.25$, and compare the denoised output obtained from our zero-shot denoising defined by \eqref{eq:one-step-denoising-super}.}
\label{figure:time_correct_factor-others}
\vspace{-0.2in}
\end{figure}

\vspace{0.05in}
\noindent
\textbf{Additional Qualitative Comparison.} Similarly to the ImageNet given in \cref{figure:time_correct_factor}, we provide more qualitative results for varying correction factor $k$ on other datasets, \textit{viz.}, 
Flower \cite{Nilsback2008flower} and Caltech \cite{FeiFei2004caltech} in \cref{figure:time_correct_factor-others}. 
Again, we confirm that using a higher correction factor enhances the clarity of denoised results, 
supporting our finding that $k$ plays a crucial role in making the text-to-image super-resolution diffusion model as effective zero-shot denoiser.

\subsection{Analysis of Inference Cost}
In this section, we evaluate our framework in terms of its inference cost from the \textsc{Predict} \cite{cohen2019certified} procedure, which requires $n$ noise samples. Specifically, we vary the sample size $n$ at the inference of our framework and compare the accuracy and inference time to observe their trade-off. 
\begin{figure}[t]
\begin{minipage}{\textwidth}
\centering
\vspace{0.08in}
\hspace{-0.2in}
\begin{minipage}[c]{0.5\textwidth}
\centering
\includegraphics[width=\linewidth]{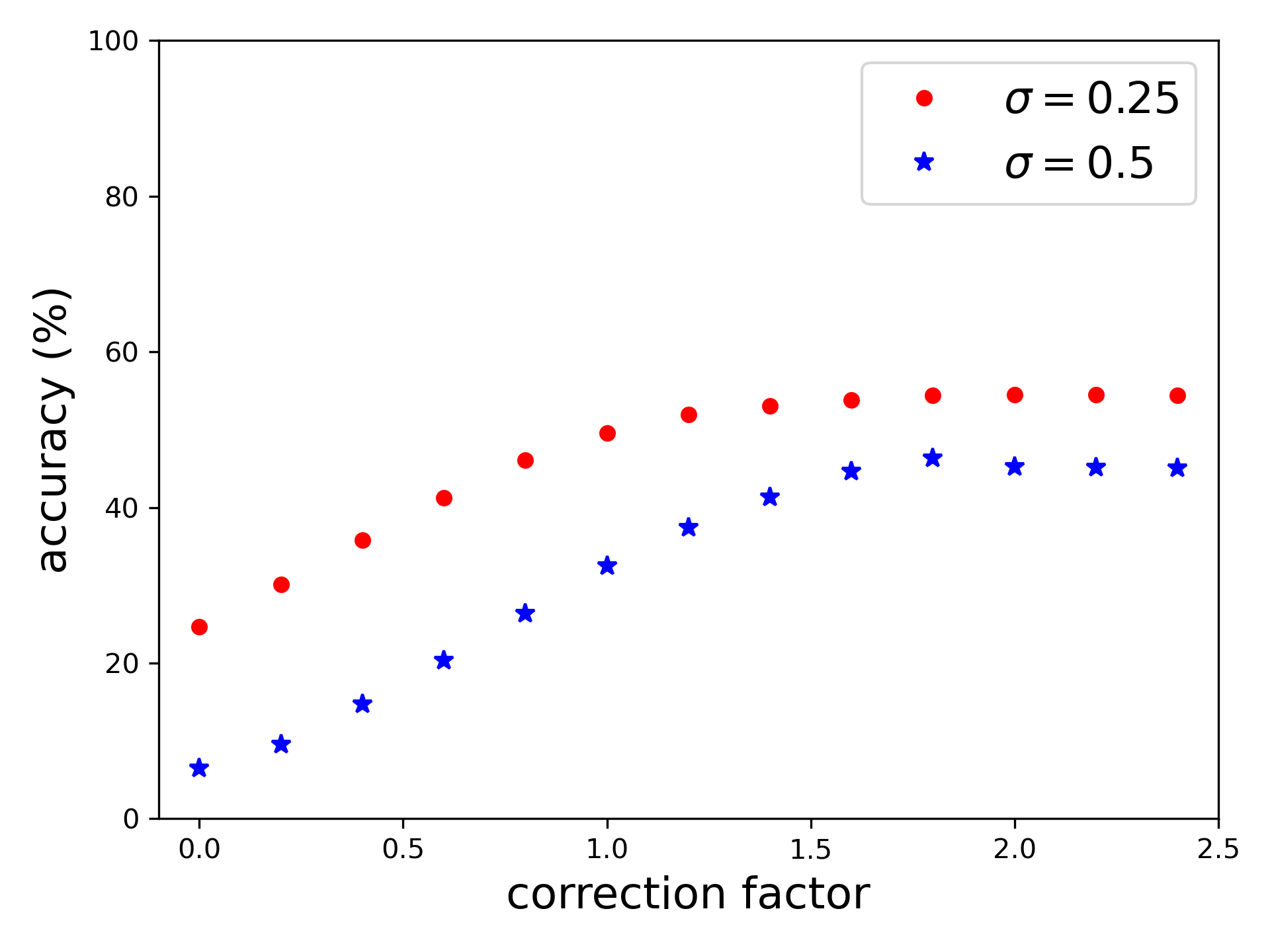}
\vspace{-0.25in}
\captionof{figure}{ImageNet accuracy (\%) on varying correction factor $k$.}\label{figure:plot_correct_factor}
\end{minipage}
\hspace{0.15in}
\begin{minipage}[c]{0.4\textwidth}
\centering
\setlength{\tabcolsep}{5pt}
\captionof{table}{Analysis of clean accuracy \vs robust accuracy at $\varepsilon=1.0$ and per-image inference time on varying $n$ for smoothing. We use a computing cluster with 4 NVIDIA A100 80GB GPUs for this experiment. Bold indicates the best. 
 }\label{table:clean-robust-table}
\vspace{0.03in}
\scalebox{0.55}{
\centering\small
\begin{tabular}{cccccc}
\toprule
\multirow{2}{*}[-1em]{}&\multicolumn{5}{c}{Sample size $n$}\\
\cmidrule{2-6}
&25&50&100&200&400\\
\midrule
Clean accuracy (\%)&\bf{58.0}&57.0&56.2&55.6&54.2\\
Robust accuracy (\%)&26.0&29.4&31.4&33.0&\bf{35.2}\\
\midrule
\shortstack{Inference time (sec) \\ \scriptsize{} }&\shortstack{\bf{0.64} \\ \scriptsize{$\pm$0.09} } &\shortstack{0.92 \\ \scriptsize{$\pm$0.10} }&\shortstack{1.39 \\ \scriptsize{$\pm$0.08} }&\shortstack{2.56 \\ \scriptsize{$\pm$0.13} }&\shortstack{5.14 \\ \scriptsize{$\pm$0.12} }\\
\bottomrule
\end{tabular}}
\end{minipage}
\vspace{-0.2in}
\end{minipage}
\end{figure}

\vspace{0.05in}
\noindent
\textbf{Trade-off between Clean and Robust.}
We analyze trade-off between clean and robust concerning the sample size $n$. To do this, we measure the \textit{empirical clean accuracy} and \textit{empirical robust accuracy} at perturbation $\varepsilon=1.0$ on ImageNet. In \cref{table:clean-robust-table}, we observe that as the sample size $n$ increases, the clean accuracy of our framework decreases, while the robust accuracy shows an upward trend. For instance, opting for $n=400$ yields a robust accuracy improvement of $9.2\%$, despite of a clean accuracy reduction of $3.8\%$, compared to the case of $n=25$. This implies that one can potentially control the clean and robust accuracy of our framework by treating $n$ as a hyperparameter. 

\vspace{0.05in}
\noindent
\textbf{Inference Time.}
We conduct an analysis on the inference time of our framework with respect to the sample size $n$. We use 4 NVIDIA A100 80GB GPUs and measure the \textsc{Predict} times of 500 images in ImageNet sequentially for computing the average. To ensure precision, we exclude the time of the first image for calculating the average, as it often includes GPU initialization time for a warm-up. In \cref{table:clean-robust-table}, we observe that increasing the sample size $n$ results in an increase in inference time. However, the table also indicates that robustness increases with a larger $n$. For example, we obtain $9.2\%$ additional robust accuracy gain by consuming $8.0\times$ inference time (\ie, $n=25 \rightarrow 400$). Therefore, the choice of an sample size $n$ also involves a trade-off between inference time and certified robust accuracy it attains.
\begin{figure}[t]
\vspace{0.1in}
\centering
\includegraphics[width=\linewidth]
{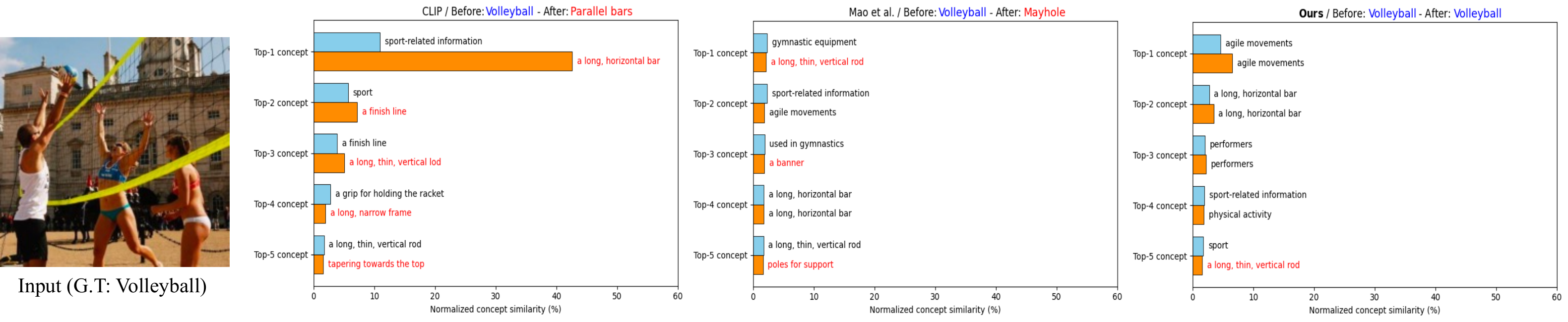}
\caption{{Comparison of the top-5 concepts with the highest similarity to an input image (labeled ``Volleyball'') before and after an $\ell_2$-adversarial attack at $\varepsilon=1.0$. Unlike other methods, our proposed framework consistently maintains relevant concepts.}}
\label{figure:concept-sim}
\vspace{-0.2in}
\end{figure}

\subsection{{Interpretablility Analysis}}

We conduct a concept-based analysis to further compare the impact of an $\ell_2$-adversary to model decisions. Specifically, we utilize the 
\emph{concept sets} originally extracted by \cite{oikarinen2023labelfree}, which comprises high-level textual descriptions corresponding to each class of ImageNet, \eg, ``marine animal'' for the ``shark'' class. For a given image $x$, we compute the \emph{(normalized) concept similarity} for each concept $c$ in a concept set $\mathcal{C}$ using the formula $\frac{\exp{sim(x, c)}}{\sum_{c'\in\mathcal{C}}\exp{sim(x, c')}}$, where
$sim(x, c)$ denotes the similarity measure between image $x$ and concept $c$ as determined by CLIP.

In \cref{figure:concept-sim}, we compare the resulting concepts for a random image labeled as ``Volleyball'' on three models: CLIP \cite{radford2021learning}, Mao et al. \cite{mao2023understanding}, and Ours applied upon CLIP. We identify and track the top-5 concepts with the highest concept similarity before and after an $\ell_2$-adversarial attack at $\varepsilon = 1.0$. 
We observe that Ours can better preserve concepts relevant to the ground truth class, ``Volleyball'', even after an attack. Compared to CLIP and Mao et al. \cite{mao2023understanding} those undergo significant changes, \eg their focus shift towards minor details (``a long, horizontal bar'' in CLIP) or unrelated concepts (``a long, thin, vertical rod'' in Mao et al. \cite{mao2023understanding}) for an image of ``Volleyball'', Ours could mostly maintain the original concepts but ``physical activity'', which is still aligned with ``Volleyball''.

\end{document}